\documentclass[journal]{IEEEtran}
\usepackage{cite}
\usepackage{ulem}
\usepackage{caption}
\usepackage{subcaption}
\usepackage{lipsum} 

\usepackage{url}
\usepackage{float}
\usepackage{booktabs} 
\usepackage{amsfonts}
\usepackage{bm}
\usepackage{algorithm}
\usepackage{algorithmic}
\usepackage{amsmath}
\usepackage{listings}
\usepackage{paralist}

\usepackage{array}
\usepackage{color}
\usepackage{tabularx}
\usepackage{longtable}
\usepackage{tabu}
\usepackage[colorlinks,linkcolor=black]{hyperref}
\usepackage{multirow}
\newcolumntype{L}[1]{>{\raggedright\let\newline\\\arraybackslash\hspace{0pt}}m{#1}}
\newcolumntype{C}[1]{>{\centering\let\newline\\\arraybackslash\hspace{0pt}}m{#1}}
\newcolumntype{R}[1]{>{\raggedleft\let\newline\\\arraybackslash\hspace{0pt}}m{#1}}

\usepackage{blkarray, bigstrut}

\usepackage{graphicx}

\usepackage{caption}

\definecolor{grey}{RGB}{130,130,130} 
\definecolor{black}{RGB}{0,0,0} 

\definecolor{red}{RGB}{255,0,0}

\makeatletter
\newcommand*{\rom}[1]{\expandafter\@slowromancap\romannumeral #1@}
\makeatother

\ifCLASSINFOpdf
\else
\fi

\begin{document}
%
\title{PECTP: Parameter-Efficient Cross-Task Prompts for Incremental Vision Transformer}
%
%
%

\author{Qian Feng,
        Hanbin Zhao\IEEEauthorrefmark{2},
        Chao Zhang,
        Jiahua Dong,
        Henghui Ding, \\
        Yu-Gang Jiang,~\IEEEmembership{Fellow, IEEE},
        and Hui Qian
        \thanks{Qian Feng, Hanbin Zhao, Chao Zhang and Hui Qian are with Zhejiang University, Hangzhou 310027, China (e-mail:  fqzju@zju.edu.cn; zhaohanbin@zju.edu.cn; zczju@zju.edu.cn; qianhui@zju.edu.cn).}
        \thanks{Jiahua Dong is with the Mohamed bin Zayed University of Artificial Intelligence, Abu Dhabi, United Arab Emirates. (e-mail: dongjiahua1995@gmail.com).}
        \thanks{Henghui Ding and Yu-Gang Jiang are with Fudan University, Shanghai, China 200433. (e-mail: hhding@fudan.edu.cn, ygj@fudan.edu.cn).}
        \thanks{The corresponding author is Hanbin Zhao.}
}


\maketitle

\begin{abstract}
  Incremental Learning (IL) aims to learn deep models on sequential tasks continually, where each new task includes a batch of new classes and deep models have no access to task-ID information at the inference time. Recent vast pre-trained models (PTMs) have achieved outstanding performance by prompt technique in practical IL without the old samples (rehearsal-free) and with a memory constraint (memory-constrained): Prompt-extending and Prompt-fixed methods. However, prompt-extending methods need a large memory buffer to maintain an ever-expanding prompt pool and meet an extra challenging prompt selection problem. Prompt-fixed methods only learn a single set of prompts on one of the incremental tasks and can not handle all the incremental tasks effectively. To achieve a good balance between the memory cost and the performance on all the tasks, we propose a Parameter-Efficient Cross-Task Prompt (PECTP) framework with Prompt Retention Module (PRM) and classifier Head Retention Module (HRM). To make the final learned prompts effective on all incremental tasks, PRM constrains the evolution of cross-task prompts' parameters from Outer Prompt Granularity and Inner Prompt Granularity. Besides, we employ HRM to inherit old knowledge in the previously learned classifier heads to facilitate the cross-task prompts' generalization ability. Extensive experiments show the effectiveness of our method. The source codes will be available at \url{https://github.com/RAIAN08/PECTP}.
\end{abstract}


%
\IEEEpeerreviewmaketitle
\section{Introduction}\label{sec:intro}
\IEEEPARstart{D}{eep} models have achieved outstanding performance in tackling a wide variety of individual machine learning tasks. However, in real-world applications, training data is often received sequentially rather than being available all at once. Therefore, equipping deep models with the ability to learn in dynamic environments is a long-term goal in Deep Learning (DL) \cite{gomes2017survey}. Incremental Learning (IL) involves dynamically learning deep models across different tasks and often suffers from performance degradation on previously learned tasks, known as catastrophic forgetting (CF) \cite{mccloskey1989catastrophic}. Recently, rehearsal-based methods can effectively mitigate forgetting in IL by keeping a few representative samples (i.e., exemplars) of old tasks in a fixed memory buffer \cite{Rebuffi_2017_CVPR,hou2019learning}. However, these approaches fail in cases with rigorous privacy concerns and severely constrained memory, where the samples of old tasks are unavailable and the memory buffer is limited. In this paper, we focus on strategies for the \uline{R}ehearsal-\uline{F}ree and \uline{M}emory-\uline{C}onstrained \uline{I}ncremental \uline{L}earning (RFMCIL), which trains the deep models without exemplars and with severe memory constraints. 

\begin{figure}
    \centering
    \includegraphics[width=1\linewidth]{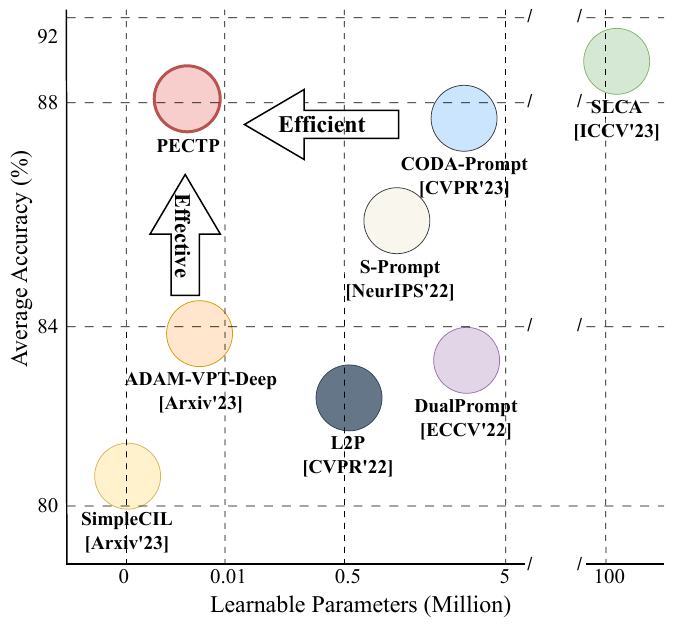}
    \caption{Comparison of different state-of-the-art incremental learning methods using Pre-Trained Models. The X-axis represents the total number of learnable parameters, and the Y-axis represents the average accuracy.} 
    \label{first_pic}

\end{figure}

The latest advances in incremental learning with Pre-Trained Models (PTMs) have already attracted widespread attention. These methods aim to address Class-Incremental Learning (CIL), a challenging setup in IL, where each new task in sequence introduces new classes while keeping the task-ID unknown during inference. Figure \ref{first_pic} showcases recent researchs on incremental learning using PTMs. SimpleCIL \cite{zhou2023revisiting} proposes directly employing the PTM for downstream incremental learning inference while adopting a prototype classifier head. SLCA \cite{zhang2023slca}, on the other hand, fully fine-tunes the PTM with different learning rates for the feature extraction network parameters and the classifier head parameters to effectively mitigate the forgetting problem. Therefore, these two methods are considered the lower bound and upper bound of incremental learning methods using PTMs, respectively. However, SimpleCIL and SLCA each face concerns regarding performance and overhead. Additionally, a major class of methods known as prompt-based Incremental Learning (PIL) has achieved great success in RFMCIL. Specifically, these methods bridge the gap between the pre-trained data and sequentially incremental tasks' data with Parameter-Efficient Fine-Tuning techniques (PEFTs), such as prompts \cite{jia2022visual}. Given the privacy concerns and memory constraints in practical IL, the tunable prompts enable the frozen PTM to adapt to different tasks effectively and efficiently \cite{Wang_2022_CVPR}.

\begin{figure}[t]
  \centering
  \begin{subfigure}[b]{0.49\columnwidth}
    \centering
    \includegraphics[width=\textwidth]{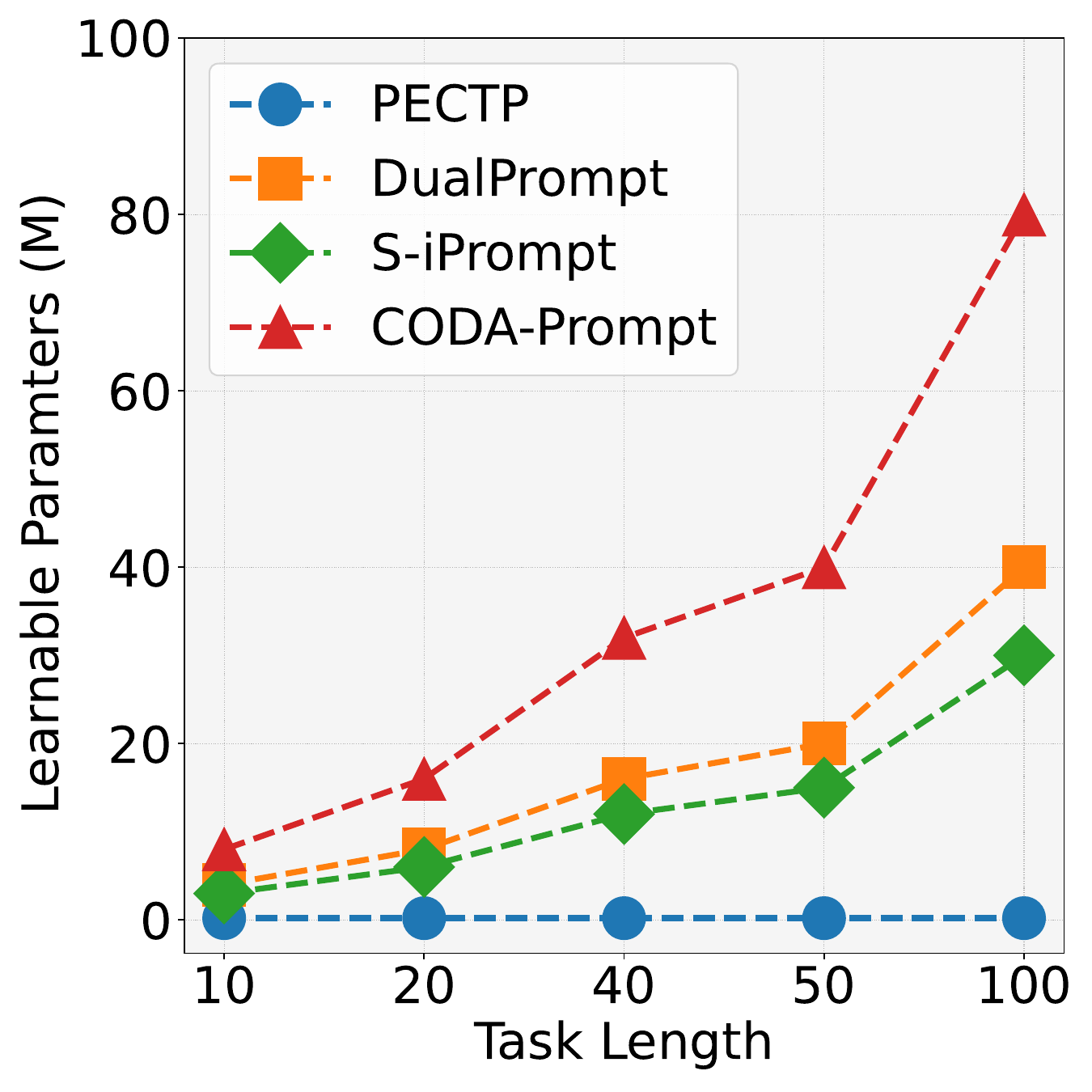}
      \vspace{-5mm}
    \caption{Overhead Problem.}
    \label{overhead_problem}
  \end{subfigure}
  \vspace{-2mm}
  \hfill
  \begin{subfigure}[b]{0.49\columnwidth}
    \centering
    \includegraphics[width=\textwidth]{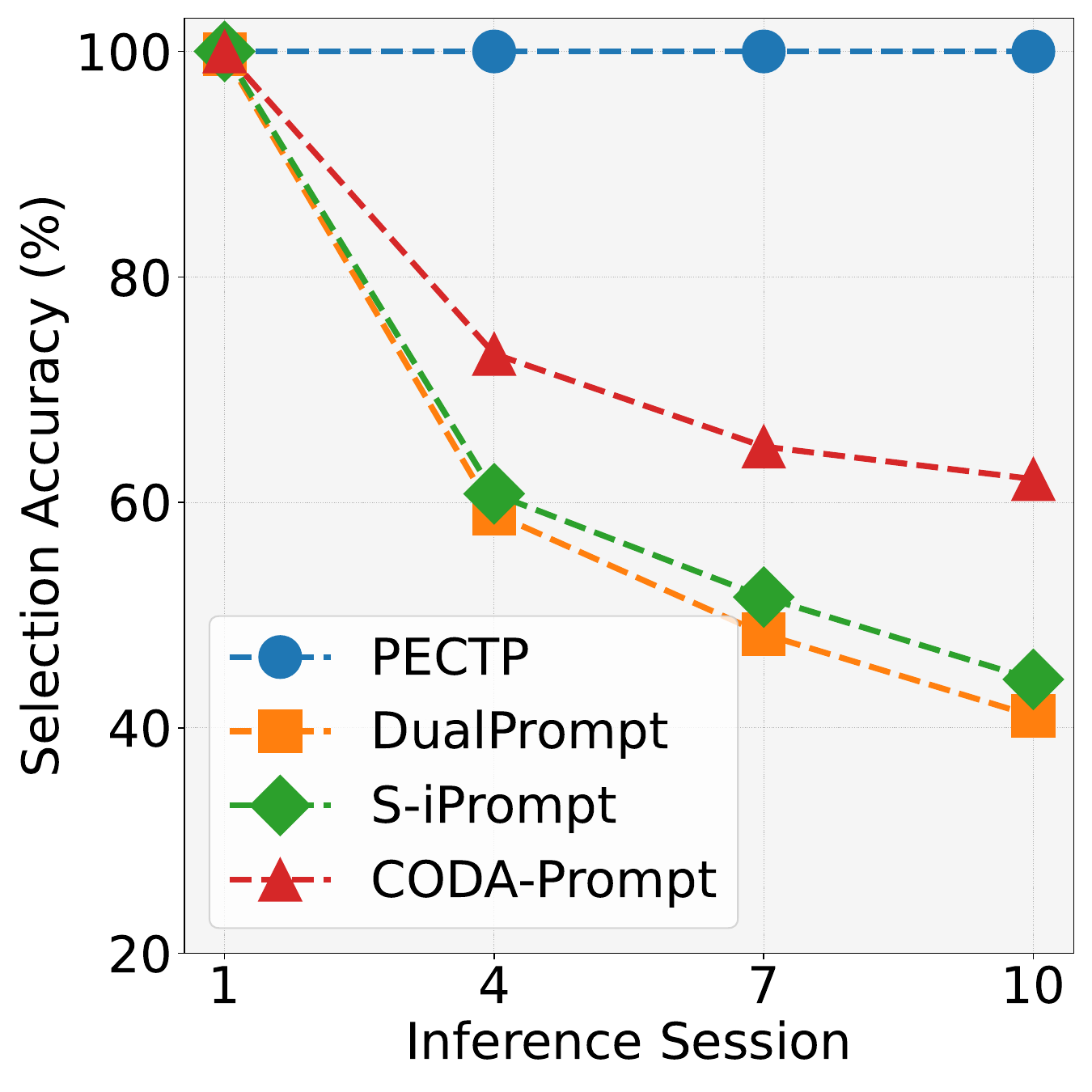}
      \vspace{-5mm}
    \caption{Selection Problem.}
    \label{2dataset_compare}
  \end{subfigure}
  \vspace{-2mm}
  \caption{Prompt-extending methods face concerns from two aspects. As incremental tasks keep increasing, the learnable parameters gradually increase (overhead problem). Simultaneously, as inference sessions progress, prompt selection accuracy becomes lower (selection problem).}
\end{figure}

Existing PIL methods mainly focus on how to utilize prompts and can be briefly separated into two categories: Prompt-extending \cite{wang2022dualprompt,wang2022s,smith2023coda} and Prompt-fixed methods \cite{zhou2023revisiting,Wang_2022_CVPR}. Prompt-extending methods need to maintain an ever-expanding prompt pool to store each task-specific set of prompts during training and select suitable set of prompts from the pool during inference. Specifically, during training, a task-specific set of prompts is newly initialized and specially trained when a new task arrives, with the aim of instructing the PTM to perform conditionally on this current incremental task. After that, this set of prompts is stored in a prompt pool that continually expands as incremental tasks are added sequentially. During inference, a prompt selection strategy is employed to first predict the task-ID to which each testing sample belongs, followed by selecting the corresponding set of prompts for further inferring. However, continually expanding the prompt pool can result in increased memory cost, which is not feasible in RFMCIL (overhead problem, as illustrated in Figure \ref{overhead_problem}). Additionally, the prompt selection strategy not only adds extra computational cost but also encounters a dilemma in modeling the relationship of prompts for different incremental tasks (selection problem, as illustrated in Figure \ref{2dataset_compare}).

Another line of work, prompt-fixed methods, learns only a single set of prompts solely for one of the entire incremental tasks (consider the learned task as the key-task), and freezes the parameters in the prompts to directly infer on the remaining incremental tasks. Although prompt-fixed methods can efficiently save memory costs, the limited knowledge of a single key-task makes it challenging for one set of prompts to guide the PTM to perform well on subsequent incremental tasks. This is particularly evident in practical IL, where incremental tasks are highly diverse, making it difficult for a single key-task to adequately represent the entire spectrum of incremental tasks.

Motivated by the above analysis, we aim to achieve an efficient and effective prompt-based method in RFMCIL. The key point is to learn a single but effective set of prompts, which not only bypasses the selection problem but also instructs the PTM to perform well on both the key-task and all incremental tasks.

In this paper, we present a \uline{P}arameter-\uline{E}fficient \uline{C}ross-\uline{T}ask \uline{P}rompt (PECTP) framework, a prompt-based method for Rehearsal-Free and Memory-Constrained Incremental Learning. PECTP aims to balance the trade-off between efficiency (parameter cost from prompts) and effectiveness (the efficacy of prompts across different incremental tasks). Our PECTP framework learns only a single but cross-task set of prompts, which are dynamically updated across all incremental tasks to continuously acquire knowledge from each incremental task and integrate it into the cross-task prompts. Specifically, we propose a Prompt Retention Module (PRM) to make these prompts effective on the learned incremental tasks. PRM restricts the evolution of cross-task prompts' parameters from two granularity: Outer Prompt Granularity (OPG) and Inner Prompt Granularity (IPG). OPG restricts parameter evolution of prompts by regularizing the output feature of prompt-based PTM. IPG constrains the variation of prompts' parameter by regularizing prompts' parameters themselves. The eventually learned single set of prompts is not only highly efficient in terms of parameter overhead but also effectively guides the PTM to perform better across various incremental tasks by integrating knowledge from each incremental task. Besides, we also propose a classifier head updating scheme named classifier Head Retention Module (HRM), which further enhances the generalization ability of cross-task prompts with the inherit knowledge from old tasks.

\begin{figure}
  \centering
  \includegraphics[width=1\linewidth]{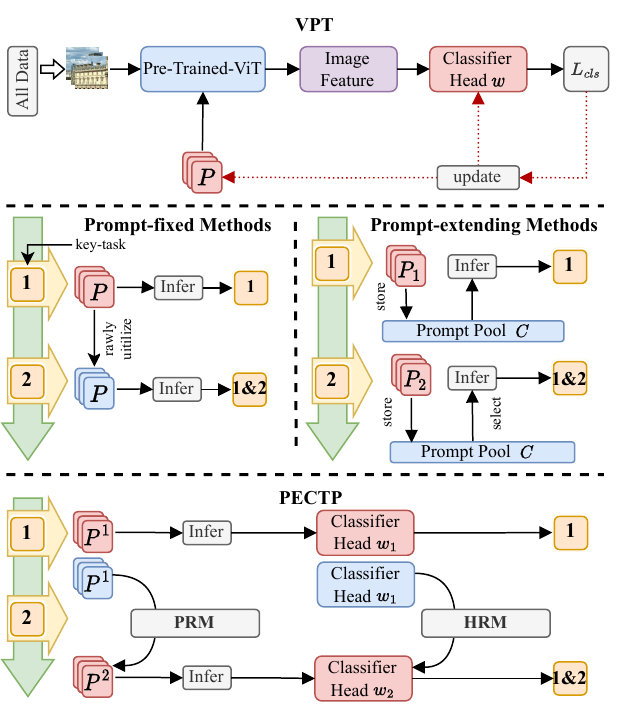}

  \caption{Illustration of sequential tasks with two incremental tasks. Prompt-fixed methods train a set of prompts $\mathcal{P}$ only on incremental task $1$, fix the parameters, and directly infer on the remaining incremental tasks. Prompt-extending methods learn a task-specific set of prompts $\mathcal{P}_{i}$ for each task. PECTP uses a single set of prompts but updates these learnable tokens on each incremental task, effectively retaining knowledge from previous tasks through two meticulously designed modules, PRM and HRM, which act on the single set of prompts and the classifier head, respectively.} 
  \label{four_compared}
  \vspace{-5mm}

\end{figure}

The core contributions of the proposed PECTP framework can be summarized as follows:

\begin{itemize}\setlength{\itemsep}{2pt}
    \item We summarize the prompt-extending and prompt-fixed IL methods among in PIL methods and propose a PECTP framework for RFMCIL, which only learns a single but cross-task set of prompts on the whole incremental tasks.
    \item We design a novel PRM to restrict the evolution of cross-task prompts to be effective on the learned incremental tasks and a HRM to inherit old knowledge to further facilitate prompts' generalization.
    \item Extensive experiments over benchmark datasets demonstrate the effectiveness of PECTP in performance and memory cost against the existing PIL methods. 
\end{itemize}
\section{Related Work}
 
\subsection{Typical Incremental Learning}

Numerous methods have been explored to enhance the ability to counteract catastrophic forgetting \cite{masana2022class,mai2022online,Dong_2023_ICCV,dong2022federated_FCIL,10323204}. These methods can be roughly divided into three main categories: $(1)$ Rehearsal-based, $(2)$ Regularization-based, and $(3)$ Architecture-based \cite{wang2024comprehensive}. Rehearsal methods \cite{isele2018selective, chaudhry2019continual, rolnick2019experience, buzzega2020dark,cha2021co2l,Rebuffi_2017_CVPR,wu2019large,ebrahimi2020adversarial,pham2021dualnet,zhao2021video,de2021continual,van2019three,wang2018progressive} explicitly retrain on a subset of exemplars while training on new tasks. These exemplars can be obtained in two ways: stored old samples from previous tasks or generated using generative models. The former faces data imbalance issues, and different methods for selecting old samples, such as herding \cite{Rebuffi_2017_CVPR, welling2009herding}, Coreset Selection \cite{yoon2021online}, and Representative Sampling \cite{zhuang2022multi}, have been proposed. The latter faces major issues related to how well the generated samples restore the original data distribution of the previous tasks, as well as the additional complexity in training the generative model, which increases overhead \cite{creswell2018generative, shin2017continual}. Regularization-based \cite{li2023variational,mazumder2022rectification,lu2024pamk,ji2022complementary,li2022ckdf,lv2024relationship,zhang2024task,wang2024layer,wei2024class,aljundi2018selfless, aljundi2017expert, farquhar2018towards, aljundi2019task} avoids storing raw inputs, prioritizing privacy, and alleviating memory requirements. Instead, these methods achieve a balance between new and old tasks by designing sophisticated but complex regularization terms. However, the soft penalty introduced from the regularization terms might not be sufficient to restrict the optimization process to stay in the feasible region of previous tasks, especially with long sequences \cite{farquhar2018towards}. Architecture-based methods \cite{pham2023continual,ji2023memorizing,kirkpatrick2017overcoming,zenke2017continual,rusu2016progressive,yoon2017lifelong,li2019learn,loo2020generalized,mallya2018packnet,serra2018overcoming,ke2020continual, fernando2017pathnet, xu2018reinforced} segregate components within the deep model for each task by expanding the network or constraining the learning rate of important parameters towards previous tasks. However, most of these methods require a task-ID during inference, which is not suitable for challenging CIL. In contrast, our method, PECTP, not only conducts inference without relying on the task-ID but also introduces a negligible number of additional parameters.

\subsection{Prompt-based Incremental Learning}
Recently, Prompt-based Incremental Learning (PIL) methods have garnered significant attention due to their utilization of PEFT techniques \cite{jia2022visual, hu2021lora} to leverage Pre-Trained Models (PTMs), achieving rehearsal-free and promising performance \cite{smith2023coda, zhou2023revisiting, wang2022s, wang2022dualprompt, zhou2022learning, Wang_2022_CVPR, zhang2023slca}. These methods follow VPT \cite{jia2022visual} to use prompts and can be typically categorized into two main types: Prompt-fixed and Prompt-extending. Prompt-fixed methods employ a single set of prompts for deliberate learning on one incremental task and keep these prompts fixed throughout subsequent tasks (e.g., ADAM-VPT-Shallow and ADAM-VPT-Deep \cite{zhou2023revisiting}). In contrast, Prompt-extending methods continually learn a novel set of prompts for each incremental task, accumulating them in an expanding prompt pool. For inference, they design different prompt selection mechanisms to predict the appropriate prompt set for each testing sample. Among them, DualPrompt \cite{wang2022dualprompt} proposed partitioning the knowledge of tasks into general and specific categories and learning them with g-prompts and e-prompts, respectively. Similarly, S-Prompt \cite{wang2022s} addressed Domain-IL by leveraging Vision-Language Models (VLMs) to further enhance the learning ability. CODA-Prompt avoids forgetting by adding new prompts, keys, and masks for each new task, while freezing previous ones and adopting the attention mechanism to allocate a subset of prompt sets for inference. 

Although the above methods show high performance, they face difficulties when applied to Rehearsal-Free and Memory-Constrained Incremental Learning (RFMCIL). Specifically, prompt-fixed methods face performance challenges. Since the introduced learnable prompts are trained only on a single incremental task, their limited representative capabilities make it difficult to provide sufficient guidance for the PTM. Meanwhile, continually growing the prompt pool and the prompt selection mechanisms in prompt-extending methods can respectively lead to increased memory overhead and extra computation cost, making them unsuitable for practical incremental learning. Conversely, PECTP not only bypasses the problem of storage and selection by maintaining only one set of prompts but also can effectively instruct the PTM across all incremental tasks. The comparison of the three types of methods is shown in Figure \ref{four_compared}.
 

\section{Prerequisite and Motivation}
\subsection{Rehearsal-Free and Memory-Constrained Incremental Learning}

Formally, Incremental Learning (IL) aims to learn a deep model on sequential tasks with novel classes. We denote the sequence of tasks as $\left\{\mathcal{T}_{i}, i=1,2,\ldots\right\}$, where $\mathcal{D}_{k} = \left\{\left(\bm{x}_{i,k},\bm{y}_{i,k}\right)\right\}_{i=1}^{n_{k}}$ is the training data corresponding to task $k$ with $n_{k}$ training samples. Here, each input sample $\bm{x}_{i,k} \in \mathbb{R}^{n}$ belongs to class $\bm{y}_{i}^{k} \in {Y}_{k}$, where ${Y}_{k}$ is the label space of task $k$. There are no overlapping classes between tasks (i.e., $Y_{k} \cap Y_{k'}=\oslash \text { if } k \neq k'$). Catastrophic forgetting problem arises because the deep model is trained only on the current task and evaluated over all the learned tasks (all encountered classes are denoted as $\mathcal{Y}_{k} = {Y}_{1} \cup, \ldots, \cup, {Y}_{k}$). In Rehearsal-Free and Memory-Constrained Incremental Learning (RFMCIL), the memory buffer is limited and samples of previous tasks can not be replayed when learning the current task. A deep image classification model is denoted as $\phi_{\theta, w}(x) = g_{w}(f_{\theta}(\bm{x}))$, where $f_{\theta}(\bm{x}):\mathcal{R}^{|\mathcal{D}^{k}|} \rightarrow \mathcal{R}^{d}$ is a feature extractor with weights $\theta$, and $g_{w}(\cdot): \mathcal{R}^{d} \rightarrow \mathcal{R}^{|\mathcal{Y}_{k}|}$ is a classifier head with weights $w$. After learning the task $k$, the goal is to learn a $\phi_{\theta,w}(\cdot)$ that can performs well on  $\mathcal{Y}_{k}$ with a memory constraint. Recent prompt-based incremental learning methods \cite{Wang_2022_CVPR} utilize a pre-trained model (PTM) with powerful representation capability, such as Vision Transformer (VIT), as the initialization for $f_{\theta}(\bm{x})$. 

\begin{figure*}
    \centering
    \includegraphics[width=1\textwidth]{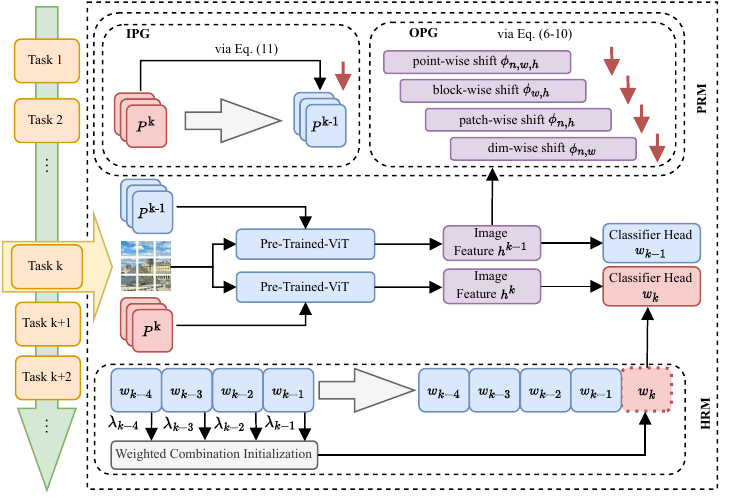}
    \caption{An architecture of the PECTP framework. When learning task $k$, a training sample is processed by both $\Phi_{k-1}$ and $\Phi_{k}$ to extract image features $\mathbf{h}^{k-1}$ and $\mathbf{h}^{k}$, respectively. In this context, $\Phi_{k-1}$ uses a fixed set of prompts $\mathcal{P}^{k-1}$, while $\Phi_{k}$ employs a learnable set of prompts $\mathcal{P}^{k}$. PRM applies constraints from the Outer Prompt Granularity on $\phi_{n,w,h}$, $\phi_{w,h}$, $\phi_{n,h}$ and $\phi_{n,w}$ (OPG). Additionally, PRM imposes constraints directly on the parameters of the single set of prompts (IPG). Moreover, HRM transfers knowledge from previously learned task-specific classifier heads to initialize the classifier head for task $k$.}
    \label{fig4}
\end{figure*}

\subsection{Pre-Trained Model with Prompt Learning}
Pre-trained models encounter domain gap problem between pre-trained data and the downstream data. Parameter-Efficient Fine-Tuning (PEFTs) technique, e.g., prompts, are proposed to address this issue, with the purpose to instruct the PTM to perform conditionally, e.g., Visual Prompt Tuning (VPT) \cite{jia2022visual}. 

Therefore, given a frozen pre-trained VIT model $f$ and a set of learnable parameters $\mathcal{P}$, namely prompts. We denote the VIT model with tunable prompts: $f + \mathcal{P}$ as $\Phi(\cdot)$, and $\mathcal{D}_\text{all}$ as the entire data of the downstream task (\textbf{Joint Training}). The objective function is a classification loss on the downstream task and defined as follows:
    \begin{align}
                \mathcal{L}_{cls} = \sum_{\left(\bm{x}, \bm{y}\right) \in \mathcal{D}_\text{all}} \mathcal{L}\left(g_{w}\left(\Phi(\bm{x})\right), \bm{y}\right),	
    \end{align}
where $\mathcal{L}_{cls}$ is a binary-cross entropy loss, $w$ stands for the parameters of classifier head and $\left[\cdot;\cdot\right]$ indicates concatenation on the sequence length dimension. Joint training is considered the upper bound for deep models in incremental learning.

\subsection{Prompt-based Incremental Learning}
\label{Prompt_extending_CIL_methods_problem}
As illustrated in Figure \ref{four_compared}, there are two main different piplines in prompt-based incremental learning methods: Prompt-extending and Prompt-fixed IL methods.

\paragraph{Prompt-extending IL methods}
Given the sequential of tasks $\left\{\mathcal{T}_{i}, i=1,2,\ldots\right\}$, prompt-extending IL methods \textbf{maintain a prompt pool} $\mathcal{C}=\left\{\mathcal{P}_{i}, i=1,2, \ldots \right\}$ during training. Before learning task $k$, a new set of prompts will be initialized: $\mathcal{P}_{k}$. The task-specific prompts $\mathcal{P}_{k}$ will be deliberately learned on task $k$ by the following loss function:
    \begin{align}
        \label{ce_loss_for_task_k}
        \mathcal{L}_{cls} = \sum_{\left(\bm{x}, \bm{y}\right) \in \mathcal{D}_{k}} \mathcal{L}\left(g_{w_{k}}\left(\Phi_{k}(\bm{x})\right), \bm{y}\right),
    \end{align}
where $f+\mathcal{P}_{k}$ is the VIT model with tunable prompt $\mathcal{P}_{k}$ and denoted as $\Phi_{k}(\cdot)$ for simplicity. $w_k$ is the classifier head corresponding to task $k$. $\mathcal{D}_{k}$ is the corresponding training data from task $k$. During inference, a selection strategy $\mathcal{F}(\cdot|\bm{x})$ is employed to select the suitable prompts $\mathcal{P}_{*}$ for each testing sample $x$:
    \begin{align}
        \mathcal{P}_{*} = \mathcal{F}\left(\mathcal{C}|\bm{x}\right).
    \end{align}
Prompt-extending IL methods raise concerns from two perspectives: (1) Increasingly expanding the capacity of $\mathcal{C}$ leads to failure in RFMCIL, and (2) The design of the prompt selection strategy $\mathcal{F}(\cdot|\bm{x})$ has a significant impact on the performance. As illustrated in Figure \ref{2dataset_compare}, we calculate the Average Selection Accuracy at each session during inference. Accuracy shapely decreases as inference session forward, which leads to poor performance.

\paragraph{Prompt-fixed IL methods}

Prompt-fixed IL methods \textbf{only learn a single set of prompts} $\mathcal{P}$ on one of the entire incremental tasks (i.e., the key-task $\mathcal{T}_{*}$, which, in the absence of prior knowledge, is typically the first incremental task \cite{zhou2023revisiting}) using the following classification loss:
    \begin{align}
        \mathcal{L}_{cls} =\sum_{\left(\bm{x}, \bm{y}\right) \in \mathcal{D}_{*}} \mathcal{L} \left(g_{w_{k}}\left(\Phi(\bm{x})\right), \bm{y}\right),
    \end{align}
where $\Phi(\cdot)$ is the VIT model with the single set of tunable prompts $\mathcal{P}$ and $w_k$ is the classifier head corresponding to task $k$. After training on the key-task $\mathcal{T}_{*}$, parameters in $\mathcal{P}$ are frozen and merged with $f$ to collectively constitute $\Phi(\cdot)$. $\Phi(\cdot)$ will be directly used in the remaining task's inference. Since the classifier head corresponding to task $i,i > 1$ has not been trained, a prototypical head approach is used to perform a non-parametric update for the classifier heads of the remaining tasks \cite{snell2017prototypical}. The only one set of prompts will not face the problem of prompt selection and remains `storage-friendly' as it does not increase additional learnable parameters with the number of tasks. Nevertheless, even if trained on the so-called `key-task', it fails to generalize to all incremental tasks. Furthermore, experimental results in Section \ref{Cross-Task Prompt vs. Key-Task Prompt} indicate that, due to the substantial difficulty of learning each incremental task, training on a single task is far from sufficient.

Based on the analysis above, we propose a simple but effective approach: \textbf{only learn a single but cross-task set of prompts}. Having only one set of prompts serves as a direct solution to bypass the prompt selection problem. Moreover, prompts are learned not solely from the key-task but encompasses all incremental tasks. Additionally, the designed PRM ensures that continuous training does not lead to the forgetting of prompts related to previous tasks. And HRM that imposed on the classifier head can further enhances cross-task prompts' generalization ability.
\begin{table*}[h!]
    \centering
    \begin{minipage}{\textwidth}
        \centering

        \begin{subtable}{\textwidth}
            \centering
            \resizebox{1\textwidth}{!}{
                \begin{tabular}{llcccccc}
                    \toprule
                    \multirow{1}*{Method} & \multicolumn{1}{c}{CIFAR Inc10} & \multicolumn{1}{c}{CUB Inc10} & \multicolumn{1}{c}{IN-R Inc10} & \multicolumn{1}{c}{IN-A Inc10} & \multicolumn{1}{c}{ObjNet Inc10} & \multicolumn{1}{c}{Omni Inc30} & \multicolumn{1}{c}{VTAB Inc10}\\
                    \hline
                    L2P & 84.33(0.41) & 56.01(0.16) & 67.20(0.41) & 38.88(0.39) & 52.19(0.40) & 64.69(1.21) & 77.10(1.01) \\
                    DualPrompt & 84.02(0.29) & 60.81(0.11) & 64.92(0.56) & 42.38(1.30) & 49.33(0.31) & 65.52(0.87) & 81.23(0.99) \\
                    ADAM-Finetune & 81.23(0.33) & 85.58(0.21) & 63.35(0.65) & 50.76(1.82) & 48.34(0.36) & 65.03(0.77) & 80.44(0.36) \\
                    ADAM-VPT-Shallow & 84.96(0.79) & 85.37(0.36) & 62.20(1.31) & 47.48(1.29) & 52.53(0.22) & 73.68(0.66) & 85.36(0.58) \\
                    ADAM-SSF & 85.14(0.52) & 85.37(0.61) & 65.00(0.84) & 51.48(1.36) & 56.64(0.56) & 74.00(1.1) & 81.92(0.74) \\
                    ADAM-Adapter & 87.49(0.98) & 85.11(0.23) & 67.20(0.86) & 49.57(1.95) & 55.24(0.64) & 74.37(1.31) & 84.35(0.91) \\
                    \hline
                    SimpleCIL & 81.26(0.20) & 85.16(0.13) & 54.55(0.16) & 49.44(0.61) & 51.13(0.33) & 73.15(0.67) & 84.38(0.31) \\
                    ADAM-VPT-Deep & 84.95(0.42) & 83.88(0.91) & 66.77(0.75) & 51.15(3.34) & 54.65(0.36) & 74.47(1.36) & 83.06(0.69) \\
                    \textbf{PECTP(Ours)} & \textbf{88.09(0.16)} & \textbf{85.69(0.52)}	& \textbf{70.28(0.19)} & \textbf{54.66(0.99)} & \textbf{58.43(0.24)} & \textbf{74.54(0.94)} & \textbf{86.32(0.70)} \\
                    \bottomrule
                \end{tabular}
                }
            \caption{Performance after the last task $\mathcal{A}_{B}$ comparison on seven datasets.}
            \label{tab:sub1}
        \end{subtable}

        \vskip\baselineskip 

        \begin{subtable}{\textwidth}
            \centering

            \resizebox{1\textwidth}{!}{
                \begin{tabular}{llcccccc}
                    \toprule
                    \multirow{1}*{Method} & \multicolumn{1}{c}{CIFAR Inc10} & \multicolumn{1}{c}{CUB Inc10} & \multicolumn{1}{c}{IN-R Inc10} & \multicolumn{1}{c}{IN-A Inc10} & \multicolumn{1}{c}{ObjNet Inc10} & \multicolumn{1}{c}{Omni Inc30} & \multicolumn{1}{c}{VTAB Inc10}\\
                    \hline
                    L2P                & 88.14(0.32) & 66.60(0.30) & 73.52(0.65) & 47.56(0.93) & 63.78(0.54) & 73.36(1.01) & 77.11(1.12) \\
                    DualPrompt           & 89.61(0.26) & 74.80(0.17) & 70.12(0.96) & 52.76(1.01) & 59.27(0.61) & 73.92(0.62) & 83.36(0.59) \\
                    ADAM-VPT-Shallow     & 89.31(0.91) & 90.15(0.96) & 70.59(1.26) & 58.11(0.67) & 64.54(1.05) & 79.63(1.13) & \textbf{87.15(0.53)} \\
                    \hline
                    SimpleCIL            & 87.13(0.31) & 90.96(0.17) & 61.99(0.25) & 60.50(0.31) & 65.45(0.32) & 79.34(0.37) & 85.99(0.11) \\
                    ADAM-VPT-Deep        & 90.29(0.22) & 89.18(0.47) & 73.76(0.76) & 62.77(3.19) & 67.83(1.69) & 81.05(1.71) & 86.59(0.73) \\
                    \textbf{PECTP(Ours)} & \textbf{92.49(0.19)} & \textbf{91.00(0.21)}	& \textbf{78.33(0.64)} & \textbf{65.74(1.22)} & \textbf{70.18(0.74)} & \textbf{81.08(0.53)} & 87.14(1.21) \\
                    \bottomrule
                \end{tabular}
            }
            \caption{Performance of $\bar{\mathcal{A}}$ comparison on seven datasets.}
            \label{tab:sub2}    
        \end{subtable}

        \caption{Main results on seven datasets. We run our benchmarks for five different shuffles of the task class order with a consistent seed (different for each trials) and report the mean and standard deviation (inside the parentheses) of these runs.}
        \label{tab:main}
    \end{minipage}
\end{table*}

\section{Prompt-based IL with Parameter-Efficient Cross-Task Prompt}

\label{section_4_begin}
	
In this section, we introduce PECTP in detail, which can enable a single set of prompts to efficiently instruct the PTM to perform effectively on the whole incremental tasks (as shown in Figure \ref{fig4}). Due to the memory-constraint in practical IL, PECTP utilizes only one set of of prompts, instead of adopting a continuously expanding prompt pool. To make these prompts generalized on the whole incremental tasks, PECTP updates the prompts on each incremental task, rather than solely on the key-task. Then, the classification loss on the current task $k$ is defined as follows:
    \begin{align}
        \mathcal{L}_{cls} = \sum_{\left(x, y\right) \in \mathcal{D}_{k}} \mathcal{L}\left(g_{w_{k}}\left(\Phi_{k}(x)\right), y\right),
    \end{align}
where $\Phi_{k}(\cdot)$ is constituted by $f$ and the single set of prompts $\mathcal{P}^{k}$ and $w_k$ is the classifier head corresponding to task $k$. We use the superscript $k$ to denote that the single set of prompts is trained on the task $k$ for simplicity. 

$\mathcal{L}_{cls}$ makes the single set of prompts effective on the current task. However, if no constraints are imposed on the parameters of $\mathcal{P}^{k}$, it will perform well on the current task $k$, but it will forget knowledge related to previous tasks, resulting in poor performance on learned tasks.

To make these prompts effective on the previous learned tasks, we propose a  Prompt Retention Module (PRM). PRM restricts the evolution of cross-task prompts' parameters from OPG (Section \ref{sec_4.2}) and IPG (Section \ref{sec_4.1}). To further improve the cross-task prompts' generalization ability, HRM is imposed on the classifier head to inherit old knowledge from learned tasks (Section \ref{sec_4.3}).

\subsection{PRM from Outer Prompt Granularity}
\label{sec_4.2}
        
Our PRM restricts parameter evolution of prompts from Outer Prompt Granularity, which regularizes the output feature of prompt-based PTM. In OPG, we introduces a set of prompt constraints, not only over the final output feature but also over the intermediate output feature of each transformer block. 

We denote each transformer block in $\Phi_{k}(\cdot)$ as $f^k_{i}, i=1,2, \ldots, N$ and the input feature of the $i$-th transformer block as $d^k_{i}$. The output of $i$-th transformer block can be formulated as follows:
    \begin{align}
        \left[\bm{c}^k_{i+1}; \bm{e}^k_{i+1}; \_\ \right]
        =
        f^k_{i}\left(
        \bm{d}^k_{i}
        \right),                     i=1,2,...,N,
        \end{align}
where $\bm{d}^k_{i} = \left[\bm{c}^k_{i}; \bm{e}^k_{i}; \bm{p}^k_{i}\right]$ is the input feature, $\bm{c}^k_{i} \in \mathbb{R}^{D}$ denotes the cls\_token, $\bm{e}^k_{i} \in \mathbb{R}^{L_{g} \times D}$ denotes the embedding of the input image with sequence length $L_{g}$ and embedding dimension $D$, $\bm{p}^k_{i} \in \mathbb{R}^{L_{p} \times D}$ denotes the prompts with prompts' length $L_{p}$. 

While learning the $k$-th incremental task, we denote the output feature of each transformer block in the model $\Phi_{k}(\cdot)$ as $\mathbf{h}_{i}^{k}=\left[\bm{c}_{i+1}^{t},\bm{e}_{i+1}^{k}, \_\ \right],i=1,2,\ldots,N$. Simultaneously, the model $\Phi_{k-1}(\cdot)$ can also extract features from each transformer block and the corresponding output features are denoted as $\mathbf{h}_{i}^{k-1},i=1,2,\ldots,N$. OPG will impose constrains on the aggregated feature, which can be formulated as $\mathbf{h}^{k} = \left[\mathbf{h}^{k}_{1}, \mathbf{h}^{k}_{2},..., \mathbf{h}^{k}_{N}\right] \in \mathbb{R}^{N \times \left(1+L_{g} + L_{p}\right) \times D}$. Each element of $\mathbf{h}^{k}$ can be denoted as $h_{n,w,h}^{k}$, where $n$ represents the block, and $w, h$ stands for patch and dimension axis, respectively. 

To approximate the prompts' parameters in $\Phi_{k}(\cdot)$ to that in $\Phi_{k-1}(\cdot)$, we aim to make the output features generated by $\Phi_{k}(\cdot)$ similar as the features generated by $\Phi_{k-1}(\cdot)$. A simple implementation is to ensure that the output features generated by both models are identical at every feature dimension (point by point). We refer to the corresponding loss as $\mathcal{L}_{\text {OPG-Point-Wise}}$:
        \begin{align}
            \mathcal{L}_{\text {OPG-Point-Wise}}=\sum_{n, w, h}\left\|\phi_{n, w, h}\right\|^{2},
        \end{align}
where $\phi_{n, w, h}=\mathbf{h}_{n, w, h}^{k-1}-\mathbf{h}_{n, w, h}^{k}$ is the point-wise representative shift between $\Phi_{k}(\cdot)$ and $\Phi_{k-1}(\cdot)$. However, part of features generated by each transformer block are weakly important or even not related to the final prediction \cite{chen2023diffrate,rao2021dynamicvit}. $\mathcal{L}_{\text {OPG-Point-Wise}}$ can make $\Phi_{k}(\cdot)$ hard to fetch the truely important features, which results in degradation in the learned tasks. Additionally, extreme constraint can disrupt the flexibility to gain novel knowledge from the current task. To address this issue, we propose a set of soft constraints on the statistic distribution of the original output features $\mathbf{h}^{k}$. $\mathbf{h}^{k}$ includes the block, patch, and dimension axis.  Then we propose to obtain the distribution knowledge of $\mathbf{h}^{k}$ from these three axes by average pooling operation as follows:

$(1)$ pooling over the block axis calculates the output feature distribution from different blocks:
        \begin{align}
            \mathcal{L}_{\text {OPG-Block-Wise}}=\sum_{w} \sum_{h} \left\|\phi_{w,h}\right\|^{2},
        \end{align}
where $\phi_{w,h}=\sum_{n}\mathbf{h}_{n, w, h}^{k-1}-\sum_{n}\mathbf{h}_{n, w, h}^{k}$ is the block-wise representative shift between $\Phi_{k}(\cdot)$ and $\Phi_{k-1}(\cdot)$.

$(2)$ pooling over the patch axis calculates the output feature distribution from different locations:
        \begin{align}
            \mathcal{L}_{\text {OPG-Patch-Wise}}= \sum_{n} \sum_{h}\left\|\phi_{n,h}\right\|^{2},
        \end{align}
where $\phi_{n,h}=\sum_{w}\mathbf{h}_{n, w, h}^{k-1}-\sum_{w}\mathbf{h}_{n, w, h}^{k}$ is the patch-wise representative shift.
   
and $(3)$ pooling over the dimension axis can calculate the output feature distribution from both the blocks and locations:
        \begin{align}
            \mathcal{L}_{\text {OPG-Dimention-Wise}}=\sum_{n} \sum_{w}\left\|\phi_{n,w}\right\|^{2}.
        \end{align}
where $\phi_{n,w}=\sum_{h}\mathbf{h}_{n, w, h}^{k-1}-\sum_{h}\mathbf{h}_{n, w, h}^{k}$ is the dimention-wise representative shift.
       
After obtaining the distribution of the original output features $\mathbf{h}^{k}$, we make the distribution information of $\Phi_{k}(\cdot)$ approximated to that of $\Phi_{k-1}(\cdot)$. Such distribution-level constraints can be considered as a form of soft constraints, effectively mitigating hard constraints that prevent the model from learning new knowledge from the current incremental task. With these OPG soft constraints, it is feasible to strike an optimal balance between learning new task knowledge and preserving old task knowledge.

\subsection{PRM from Inner Prompt Granularity}
\label{sec_4.1}

Our PRM constrains prompt parameter variation from Inner Prompt Granularity. Specifically, while learning the $k$-th incremental task, the prompts in $\Phi_{k}(\cdot)$ should keep the knowledge of the learned prompts in $\Phi_{k-1}(\cdot)$ and obtain the knowledge of current task $k$. In order to make $\mathcal{P}^{k}$ effective on task $k-1$, we impose a Inner Prompt Granularity Loss $\mathcal{L}_{\text {IPG}}$ between $\mathcal{P}^{k}$ and $\mathcal{P}^{k-1}$:
        \begin{align}
            \label{ipg_label}
            \mathcal{L}_{\text {IPG}}=\sum_{n} \sum_{w} \sum_{h}\left\|{p}_{n, w, h}^{k-1}-{p}_{n, w, h}^{k}\right\|^{2},
        \end{align}
where the total prompts can be formulated as $\mathcal{P}^{k}=\left[\bm{p}_{1}^{k},\bm{p}_{2}^{k},\ldots,\bm{p}_{N}^{k} \right] \in \mathbb{R}^{N \times L_{p} \times D}$. Each element of $\mathcal{P}^{k}$ can be denoted as $p_{n,w,h}^{k}$.

\subsection{HRM on the Classifier Head}
\label{sec_4.3}

The features extracted through the VIT need to be mapped to the classification space through a classifier head, as shown below:
            \begin{align}
            \hat{y} = g_{w}(\Phi(\bm{x})),
            \end{align}
where $w$ represents the parameters of the classifier head.

Prompt-based IL methods typically learn a unified classifier head $w$. As incremental tasks are added, $w$ will continuously expand. Therefore, to further enhance the generalization ability of the cross-task prompts, we propose a classifier head Retention Module. Specifically, we partition the parameters $w$ of the classifier head trained on task $k-1$ into $w_{1}$, $w_{2}$, ..., $w_{k-1}$. Here, $w_{i}$ represents the parameters of the classifier head for task $i$ and its corresponding information. The previous $w_{1}$, $w_{2}$, ..., $w_{k-1}$ are frozen to avoid catastrophic forgetting. While learning task $k$, to effectively retain knowledge from previous tasks, the current classifier head is initialized with a weighted combination of all previous classifier heads:
    \begin{align}
        \label{Classification_layer_transfer}
        w_{k} \leftarrow \frac{1}{k-1}\sum_{i=1}^{k-1}\gamma_{i}w_{i},
    \end{align}
where $\gamma_{i}$ is a hyperparameter that controls the strength of inherited old knowledge from each task to facilitate $w_{k}$ in learning the current task.

\subsection{Full Optimization}

Our model is trained with three parts of losses: $(1)$ the classification loss $\mathcal{L}_{cls}$, a binary-cross entropy to learn on the current incremental task, $(2)$ an Inner Prompt Granularity loss $\mathcal{L}_{\text {IPG}}$ in PRM to regularize the prompts' parameter themselves, and $(3)$ an Outer Prompt Granularity loss $\mathcal{L}_{\text {OPG}}$ in PRM to restrict parameter evolution of prompts by regularizing the output feature of prompt-based PTM. The total loss is:
    \begin{align}
    \label{all_loss_func}
        \mathcal{L}_{\text{all}} = \mathcal{L}_{cls} + \alpha \mathcal{L}_{\text {IPG}} + \beta \mathcal{L}_{\text{OPG}},
    \end{align}        
where $\alpha$ and $\beta$ are two hyperparameters to maintain the balance between learning new task knowledge and preserving old task knowledge.

We provide the pseudo-code of PECTP in Algorithm \ref{algorithm_1}. When learning the task $1$, PECTP will degrade to ADAM-VPT-Deep, which guarantees the performance lower bound. Unlike ADAM-VPT-Deep, PECTP will train the single set of prompts on each incremental task from task $2$ to task $m$, thereby continually accumulating knowledge from new tasks. The loss functions $\mathcal{L}_{\text {IPG}}$ and $\mathcal{L}_{\text{OPG}}$ balance the trade-off between preserving old knowledge and accumulating novel knowledge.

\begin{table}[t]
    \centering
        \scriptsize
        \small
        \setlength{\tabcolsep}{1.5pt}
        \begin{tabular}{lcccc}
            \toprule
            \multirow{2}{*}{Task Sequence Length} & \multicolumn{4}{c}{IM-R} \\
            ~ & 20(Inc10) & 40(Inc5) & 50(Inc4) & 100(Inc2) \\
            \hline
            \addlinespace 
            L2P                   & 65.86  & 59.22 & 57.28 & 35.56 \\
            DualPrompt              & 67.87  & 55.22 & 58.61 & 39.66 \\
            SimpleCIL               & 54.55  & 54.55 & 54.55 & 54.55 \\
            ADAM-VPT-Deep(Baseline) & 66.47  & 64.3  & 60.35 & 54.07 \\
            \textbf{PECTP(Ours)}    & \textbf{70.01}  & \textbf{68.15} & \textbf{66.18} & \textbf{59.75} \\
            \bottomrule

        \end{tabular}
        
    \caption{Results of $\mathcal{A}_{b}$ in long-sequence learning tasks on IM-R Inc10, Inc5, Inc4 and Inc2.}
    \label{Pressure_Test}
\end{table}%

\begin{algorithm}[t]
    \caption{Our PECTP Framework to PIL}
    \label{alg:algorithm}
    \textbf{Input}: Sequential supervised tasks $\mathcal{T}_1, \mathcal{T}_2,...,\mathcal{T}_m$ and corresponing dataset $\mathcal{D}_1, \mathcal{D}_2,...,\mathcal{D}_m$, Initialized a single set of Prompts $\mathcal{P}$ and Initialized a Classifier Head $W$; Pre-trained VIT $f$, Hyperparameters $\alpha$, $\beta$ and $\gamma$;\\
    \textbf{Output}: Prompts parameter $\mathcal{P}^*$ and Classifier Head parameter $W^*$;
	\begin{algorithmic}[1] 
         \FOR{$i=1,...,m$}
		\IF {$i==1$}
            \STATE \text{Get} $\Phi_{i}$ \text{by merging}$ {\mathcal{P}}_i$ \text{with} Pre-trained VIT $f$;
            \STATE \text{Get} $\mathcal{L}_{cls}$ \text{by feeding each training sample from} $\mathcal{D}_{i}$ into $\Phi_{i}$;
		\STATE ${\mathcal{P}}_i, w_{i}\gets \arg\min\limits_{\mathcal{P}, w}\mathcal{L}_{\text{all}}$ via Eq.~\eqref{ce_loss_for_task_k};
		\ELSE
            \STATE \text{Get} $w_{i}$ \text{by initializing with} $\left\{w_{j}, j < i\right\}$ via Eq.~\eqref{Classification_layer_transfer};
            \STATE \text{Get} $\Phi_{i}$ \text{by merging learnable} ${\mathcal{P}}^i$ \text{with} Pre-trained VIT $f$; 
            \STATE \text{Get} $\Phi_{i-1}$ \text{by merging fixed} ${\mathcal{P}}^{i-1}$ \text{with} Pre-trained VIT $f$;
            \STATE \text{Get} $\mathcal{L}_{\text{OPG}}$ \text{by feeding each training sample from} $\mathcal{D}_{i}$ into $\Phi_{i}$ and $\Phi_{i-1}$;
            \STATE \text{Get} $\mathcal{L}_{\text{IPG}}$ via Eq.~\eqref{ipg_label};            

		\STATE ${\mathcal{P}}_i, w_{i}\gets \arg\min\limits_{\mathcal{P}, w}\mathcal{L}_{\text{all}}$ via Eq.~\eqref{all_loss_func};
		\ENDIF
		\STATE \textbf{return} $\mathcal{P}^*=\mathcal{P}^{m}$ \textbf{and} $\mathbf{W}^*=\left[w_1;w_2;...;w_m\right]$.\\
	    \ENDFOR
	\end{algorithmic}
    \label{algorithm_1}
\end{algorithm}

\section{Experimental Results}
\label{others}	
\subsection{Experimental Setup}
\label{section_5.1}

\paragraph{Datasets} We follow \cite{zhou2023revisiting} and conduct the experiments on seven datasets: CIFAR100 \cite{krizhevsky2009learning} (CIFAR), CUB200 \cite{wah2011caltech} (CUB), ImageNet-R \cite{hendrycks2021many} (IN-R), ImageNet-A \cite{hendrycks2021natural} (IN-A), ObjectNet \cite{barbu2019objectnet} (ObjNet), Omnibenchmark \cite{zhang2022benchmarking} (Omni) and VTAB \cite{zhai2019large}. As described in \cite{zhou2023revisiting}, the last four datasets have a large domain gap with the pre-trained dataset ImageNet. IM-A and ObjNet include the challenging samples that PTMs with ImageNet can merely handle, while Omni and VTAB contain diverse classes from multiple complex realms. 

\paragraph{Training Details} 
For compared methods, we adopt the implementation\footnote{https://github.com/JH-LEE-KR/l2p-pytorch} of L2P \cite{Wang_2022_CVPR} and the implementation\footnote{https://github.com/JH-LEE-KR/dualprompt-pytorch} of DualPrompt \cite{wang2022dualprompt}. We follow the implementations in ADAM\footnote{https://github.com/zhoudw-zdw/RevisitingCIL} \cite{zhou2023revisiting} to re-implement other compared methods with VIT, i.e., ADAM-Finetune, ADAM-VPT-Shallow, ADAM-VPT-Deep, ADAM-SSF and ADAM-Adapter. Our PECTP is based on the ADAM-VPT-Deep and utilize the same hyper-parameters in \cite{zhou2023revisiting} (e.g., learning rate, epoch, weight decay, the number of prompts). We utilize the PTM VIT-B/16-IN21K, which is pre-trained on ImageNet21K. Following \cite{Wang_2022_CVPR}, we use the same data augmentation for all methods, i.e., random resized crop and horizontal flip. Input images are resized to $224 \times 224$ before feeding into the model. Following \cite{Rebuffi_2017_CVPR}, all classes are randomly shuffled with Numpy random seed 1993 before splitting into incremental tasks. Furthermore, PRM in PECTP uses the hyperparameter $\alpha$ and $\beta$ to maintain the balance between learning new task knowledge and preserving old task knowledge. We adopt the hyperparameters $\alpha=\frac{1}{3.5\times10^5}, \beta=\frac{1}{4\times10^2}$ for CIFAR, $\alpha=\frac{1}{8\times10^3}, \beta=\frac{1}{5\times10^2}$ for CUB, $\alpha=\frac{1}{4.5\times10^4}, \beta=\frac{1}{5\times10^2}$ for IM-R, $\alpha=\frac{1}{2\times10^4}, \beta=\frac{1}{5\times10^2}$ for IM-A, $\alpha=\frac{1}{2\times10^4}, \beta=\frac{1}{2\times10^2}$ for ObjNet, $\alpha=1/\frac{1}{1.5\times10^4}, \beta=\frac{1}{1\times10^2}$ for Omni and $\alpha=\frac{1}{9\times10^4}, \beta=\frac{1}{2\times10^2}$ for VTAB. 

    \begin{table}[t]
        \centering
    
        \resizebox{1\linewidth}{!}{
            \huge
            \begin{tabular}{lcccccccc}
                \toprule
                \multirow{2}{*}{Method} & \multicolumn{2}{c}{CIFAR Inc10} & \multicolumn{2}{c}{IN-A Inc10} & \multicolumn{2}{c}{CUB Inc10} & \multicolumn{2}{c}{IN-R Inc10}  \\
                ~ & $\bar{\mathcal{A}}$ & $\mathcal{A}_{B}$ & $\bar{\mathcal{A}}$ & $\mathcal{A}_{B}$ & $\bar{\mathcal{A}}$ & $\mathcal{A}_{B}$ & $\bar{\mathcal{A}}$ & $\mathcal{A}_{B}$ \\
                \hline
                \addlinespace 
    
                PlainCIL & 90.68 & 87.00 & 59.95 & 48.44 & 84.30 & 77.27 & 68.40 & 61.28 \\
                SimpleCIL & 87.13 & 81.26 & 60.50 & 49.44 & 90.96 & 85.16 & 61.99  & 54.55 \\
                Baseline &  90.19  & 84.66 & 60.59 & 48.72 & 89.48 & 83.42 & 74.46 & 66.47 \\
                
                Baseline+OPG & 92.43 & 87.66 & 65.48 & 53.92 & 90.01 & 85.09 & 77.03 & 69.38 \\
                Baseline+IPG & 91.70 & 87.60 & 61.41 & 49.11 & 89.69 & 84.01 & 75.91 & 67.43 \\
                Baseline+IPG+OPG &\textbf{92.59} & \textbf{87.73} & \textbf{66.21} & \textbf{55.43} & \textbf{91.01} & \textbf{85.11} & \textbf{77.42} & \textbf{70.01} \\
                \bottomrule
    
            \end{tabular}}    
        \caption{Results of ablating components of PRM (i.e., OPG, IPG) on 4 datasets.}
        \label{components_in_PRM}
    \end{table}%
    
    \begin{table}[t]
        \centering
        \small
        \setlength{\tabcolsep}{9pt}
        \begin{tabular}{l*{4}{c}}
            \toprule
            \multirow{2}{*}{Initialization Methods} & \multicolumn{2}{c}{CIFAR Inc10} & \multicolumn{2}{c}{IM-R Inc10} \\
            & $\bar{\mathcal{A}}$ & $\mathcal{A}_{B}$ & $\bar{\mathcal{A}}$ & $\mathcal{A}_{B}$ \\
            \midrule
            zero-Init   & 87.30  & 81.55 & 70.06 & 61.01 \\
            uniform-Init      & 91.86  & 87.50 & 76.10  & 68.13 \\
            kaming-Init & 92.27  & 87.69 & 76.33  & 68.41 \\
            \textbf{old-Init(Ours)}  & \textbf{92.49} & \textbf{88.09} & \textbf{78.33} & \textbf{70.28}  \\
            \bottomrule
        \end{tabular}
        \caption{Influence of different implementations in the classifier head on CIFAR and IM-R.}
        \label{paper_xxxtable1}
        \vspace{-4mm}
    \end{table}

\paragraph{Evaluation Protocol} We adopt two metrics for incremental learning, including the final average accuracy ($\mathcal{A}_{B}$) and cumulative average accuracy ($\bar{\mathcal{A}}$) to evaluate the performance on all seen classes after learning each new incremental task.  Specifically, we define the accuracy on the $i$-th task after learning the $t$-th task as $A_{i,t}$, and define the Final Average Accuracy $\mathcal{A}_{B}$ as follows:
\begin{align}
\mathcal{A}_{B} = \frac{1}{B} \sum_{i=1}^{B} {A_{i,B}}.
\end{align}
Likewise, the Cumulative Average Accuracy $\bar{\mathcal{A}}$ is defined as follows:
\begin{align}
    \bar{\mathcal{A}} = \frac{1}{B} \sum_{b=1}^{B} \mathcal{A}_{b}.
\end{align}
$\mathcal{A}_{B}$ is the primary metric to evaluate the final performance of incremental learning, $\bar{\mathcal{A}}$ further reflects the historical performance. We further provided the results of $\bar{\mathcal{A}}$ with mean and standard deviation (inside the parentheses) on seven benchmarks to validate the effectiveness of our methods. Besides, as illustrated in Figure \ref{2dataset_compare}, we also calculated the prompt selection accuracy after training on each incremental task by comparing the set of selected prompts from prompt-extending methods with the ground-truth set of prompts.

\begin{figure*}[htbp]
  \centering
  \begin{subfigure}[b]{0.49\textwidth}
    \centering
    \includegraphics[width=\textwidth]{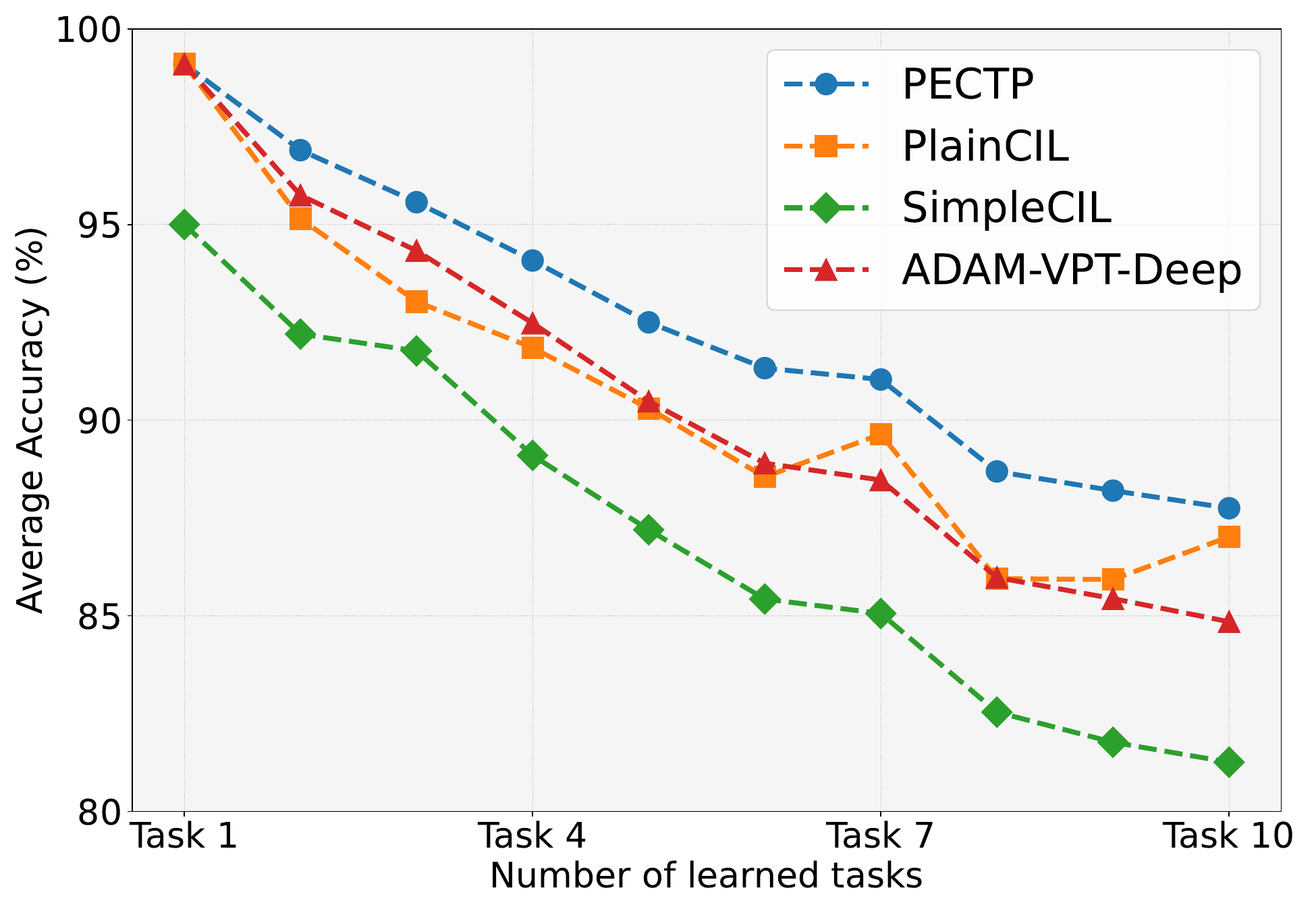}
    \caption{CIFAR Inc10.}
    \label{gain_on_each_cifar}
  \end{subfigure}
  \hfill
  \begin{subfigure}[b]{0.49\textwidth}
    \centering
    \includegraphics[width=\textwidth]{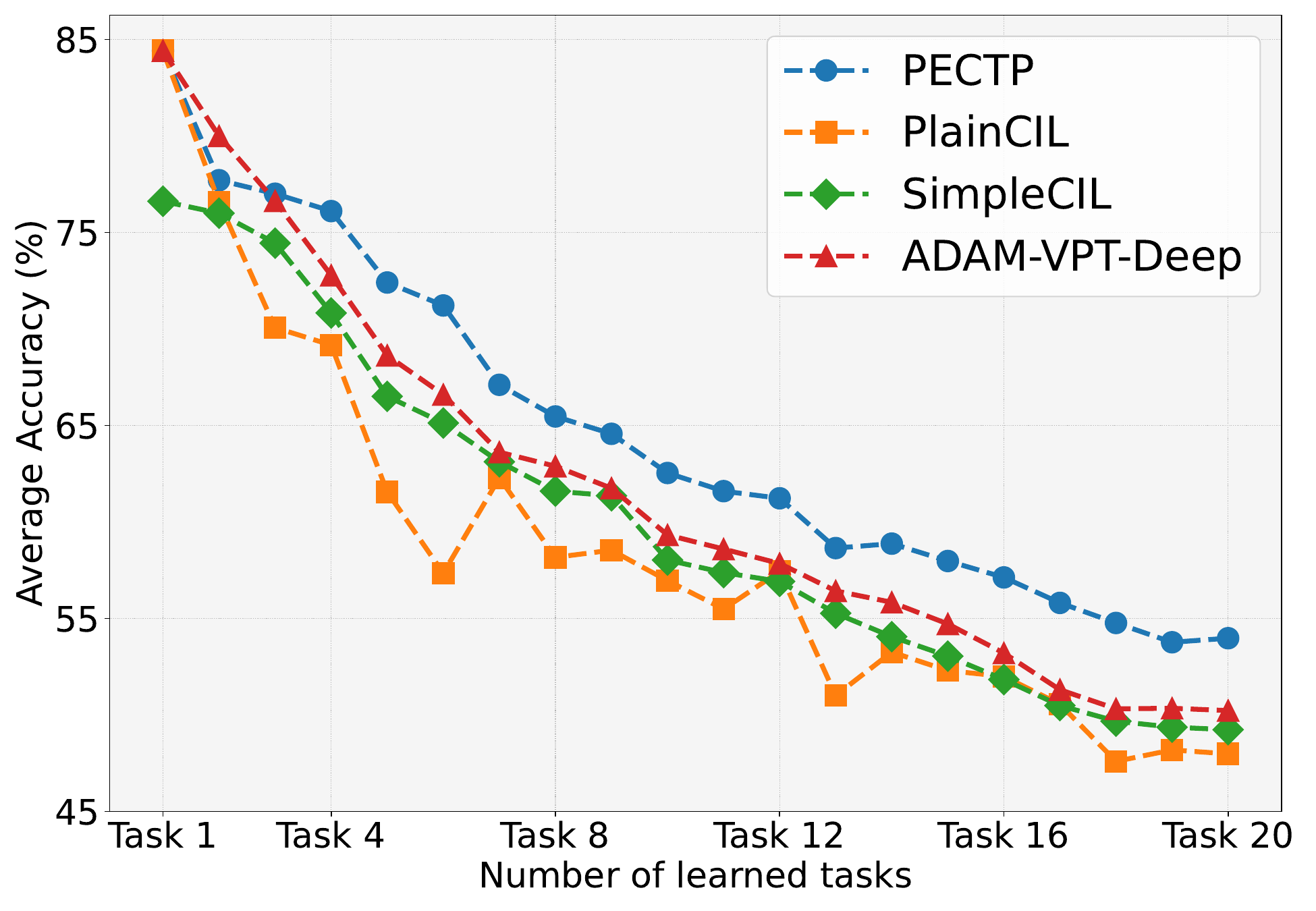}
    \caption{IM-A Inc10.}
    \label{gain_on_each_ima}
  \end{subfigure}
  \caption{The detailed improvement of PECTP over the baseline for each task. The x-axis denotes each incremental task. Apart from ADAM-VPT-Deep, we also present the results of PlainCIL and SimpleCIL.}
\end{figure*}

\begin{figure*}[h]

  \centering
  \begin{subfigure}[b]{0.32\textwidth}
    \centering
    \includegraphics[width=\textwidth]{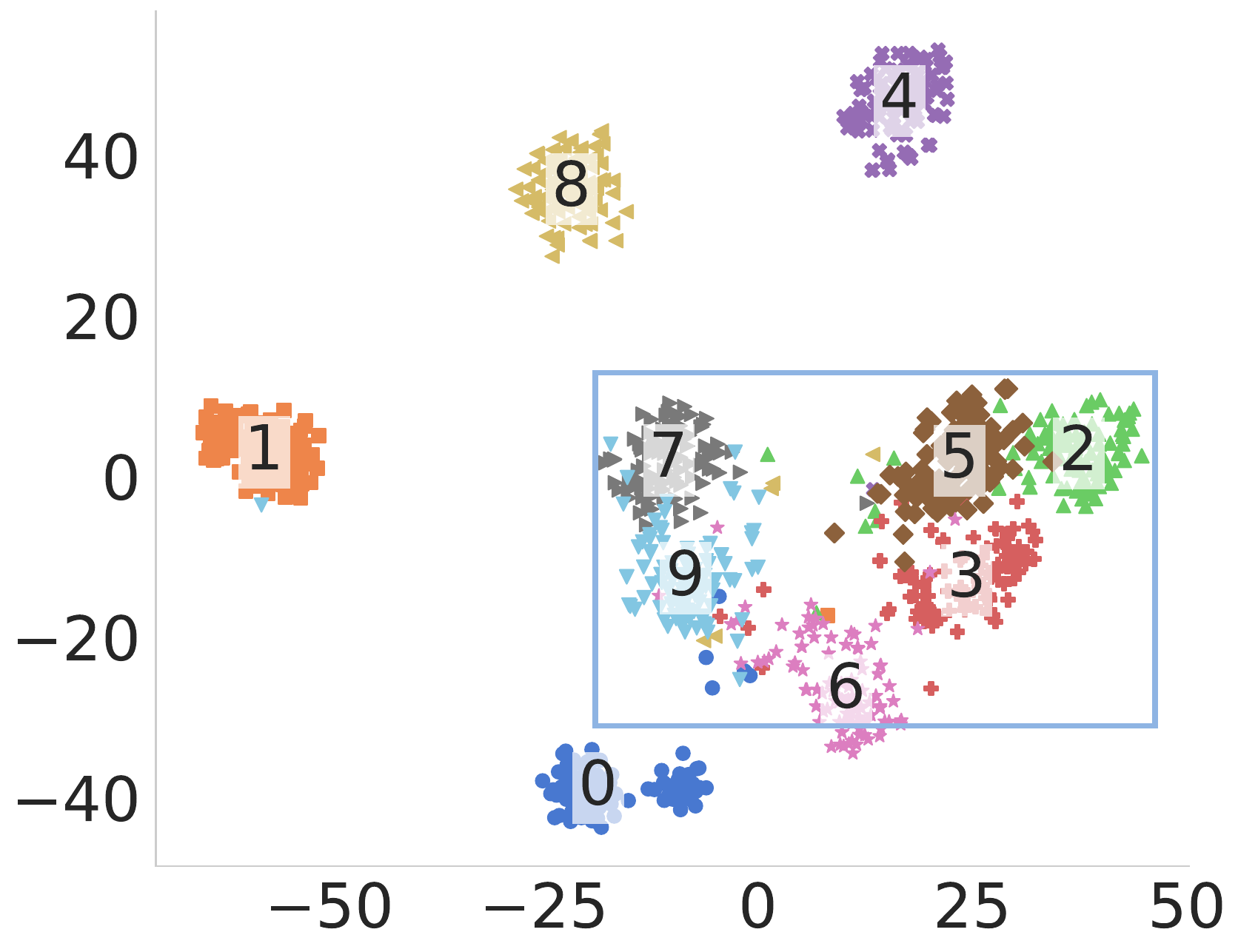}
    \caption{Prompt-fixed methods.}
    \label{fig:sub1}
  \end{subfigure}
  \hfill
  \begin{subfigure}[b]{0.32\textwidth}
    \centering
    \includegraphics[width=\textwidth]{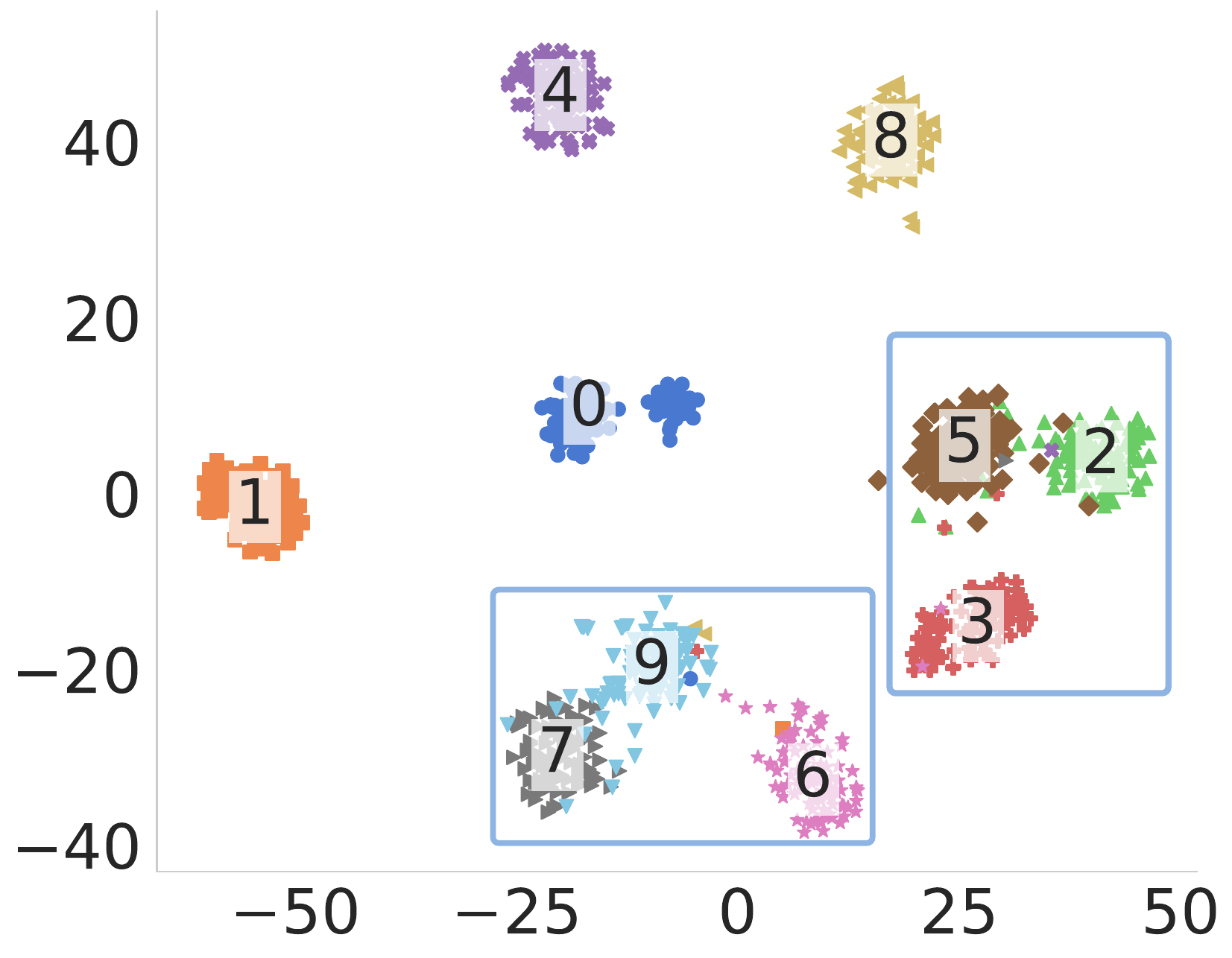}
    \caption{Prompt-extending methods.}
    \label{fig:sub2}
  \end{subfigure}
  \hfill
  \begin{subfigure}[b]{0.32\textwidth}
    \centering
    \includegraphics[width=\textwidth]{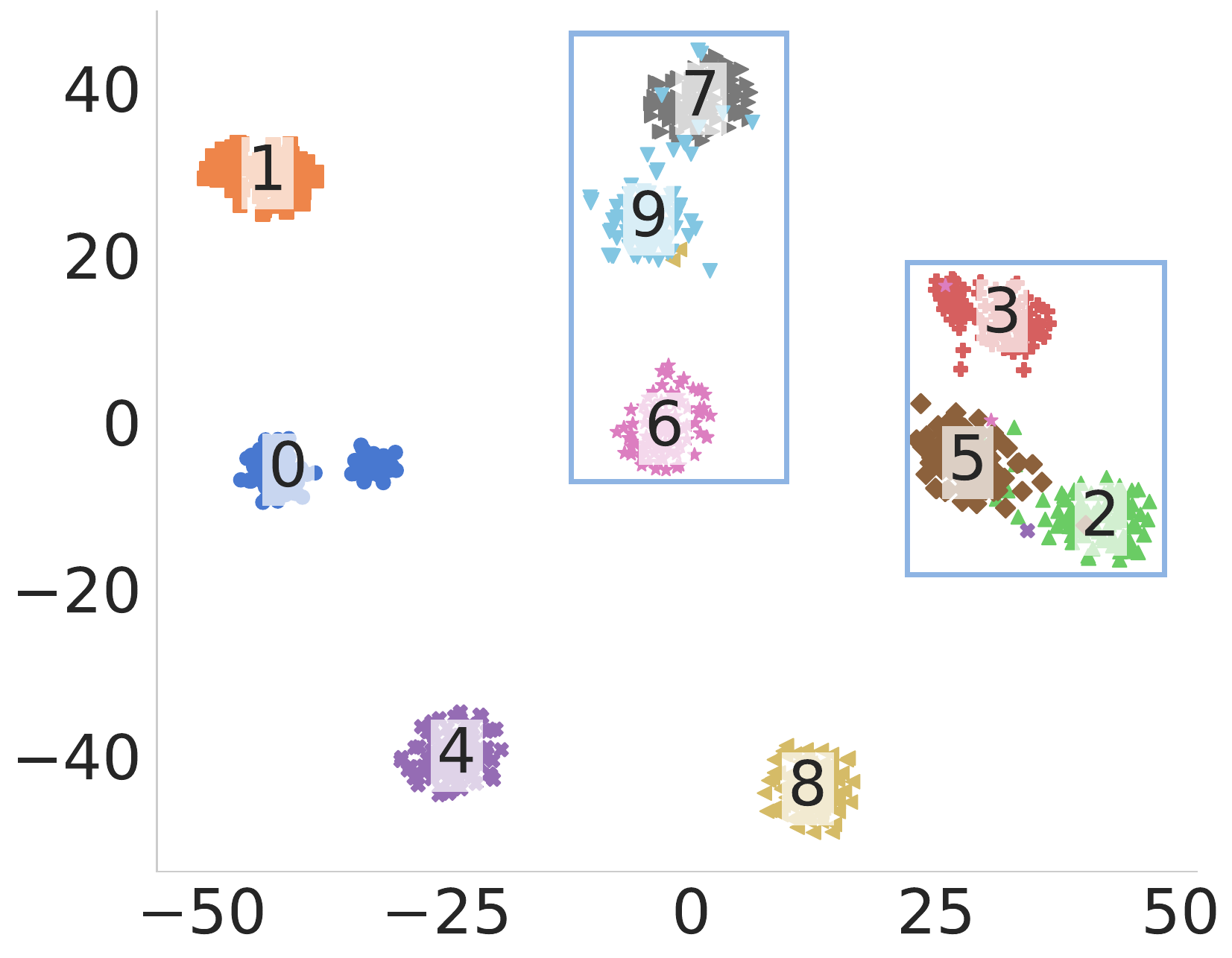}
    \caption{PECTP.}
    \label{fig:sub3}
  \end{subfigure}

\caption{T-SNE visualization of features obtained by prompt-fixed methods, prompt-extending methods and PECTP on each class from task $10$ in CIFAR.} 
\label{paper:tsne}
\end{figure*}




\subsection{Comparison to Previous Methods}
\paragraph{Main Results}

We compare our method with recent prompt-based IL methods: L2P \cite{Wang_2022_CVPR}, DualPrompt \cite{Wang_2022_CVPR}, CODA-Prompt \cite{smith2023coda}, SimpleCIL \cite{zhou2023revisiting}, and ADAM \cite{zhou2023revisiting}. ADAM has different variants with various adaptation techniques (i.e., ADAM-Finetune, ADAM-VPT-Shallow, ADAM-VPT-Deep, ADAM-SSF, ADAM-Adapter). DualPrompt and CODA-Prompt are Prompt-extending methods. L2P, ADAM-VPT-Deep, ADAM-VPT-Shallow and Our proposed PECTP are Prompt-fixed methods.

We show the performance of these methods on seven datasets in Table \ref{tab:sub1} and Table \ref{tab:sub2}. On typical IL datasets CIFAR, CUB and IM-R, our method achieves 88.09\%, 84.69\% and 70.28\% on $\mathcal{A}_{B}$, outperforming the baseline ADAM-VPT-Deep by 3.14\%, 1.81\% and 3.51\%, respectively. On the challenging IM-A and ObjNet datasets, our method achieves 54.66\% and 58.43\% on $\mathcal{A}_{B}$, surpassing ADAM-Adapter by 5.09\% and 3.19\%, respectively. On Omni and VTAB datasets, the $\mathcal{A}_{B}$ of our method is 9.02\% and 5.00\% higher than that of DualPrompt respectively. 

\paragraph{Pressure Test}
Learning in the context of long sequential tasks has long been regarded as a more challenging setting in IL. As illustrated in Figure \ref{overhead_problem} and Figure \ref{2dataset_compare}, Prompt-extending methods face not only storage difficulties during incremental sessions, which become extremely long, but also hardships in prompt selection. In contrast, PECTP only maintain a single set of prompts, which can bypass the prompt selection problem. Results in Table \ref{Pressure_Test} indicate that, although PECTP has fewer learnable parameters compared to prompt-extending methods, it can still perform well in some extreme tasks (long-sequence learning tasks).

\subsection{Ablation Experiments}
\label{ablats}
\paragraph{Effect of PRM} As described Section \ref{section_5.1}, we select the ADAM-VPT-Deep \cite{zhou2023revisiting} as the Baseline of PECTP. We have conducted the experiments to validate the effectiveness of our PRM (OPG and IPG). As shown in Table \ref{components_in_PRM}, both IPG and OPG can improve the performance of Baseline on CIFAR, CUB, IM-R and IM-A datasets. While utilizing OPG and IPG simultaneously, our method achieves the highest performance, illustrating that these two granularity are necessary and complementary for keeping the old task knowledge of prompts. Besides, we also add another baseline, denoted as PlainCIL, which has no constrains on the prompt while training on the current task with $\mathcal{L}_{cls}$. As mentioned in the Section \ref{section_4_begin}, prompt will be prone to forget the knowledge related to previous tasks. Therefore, PlainCIL performs even worse than SimpleCIL.     


\paragraph{Effect of HRM} HRM initializes the current classifier head with a weighted combination of all previous classifier heads. We explore the experiment results with different initialization methods: (1) initialized with zeros (zero-Init), (2) initialized with the previously learned classifier heads (old-Init) which is the default settings in PECTP, (3) initialized with Uniform-distribution (uniform-Init) and (4) initialized with Kaming (kaming-Init). The results shown in Table \ref{paper_xxxtable1} indicate that the classifier head also encapsulates the relationship between old and new tasks. The different initialization methods of the classifier head significantly impact the performance. The proposed HRM in PECTP effectively transfers knowledge relevant to new tasks from old tasks and facilitates learning on new tasks.

\subsection{Overheads}
To validate the computational efficiency of our PECTP framework, we select four metrics to compare PECTP with other baseline methods: Prompt Number, Learnable Parameters, Training Time, and Selection Time.
    \begin{table*}[h]
        \centering

        \resizebox{1\linewidth}{!}{
            \small
            \begin{tabular}{llcccccc}
                \toprule
                \multicolumn{2}{c}{\multirow{2}{*}{Method}} & \multicolumn{1}{c}{Performance} & \multicolumn{5}{c}{Overheads} \\
                & & $\mathcal{A}_{B}$ & PN & LP(M) & Flops(G) & TT(s) & ST(s)  \\ 
                \hline
                \addlinespace 
                \multirow{2}{*}{Prompt-Extending} 
                & DualPrompt & 83.05 & 605 & 2.39(44.25$\times$) & 35.19(2.03$\times$) & 26.06$\pm$0.40 & 1.37$\pm$0.16 \\
                & CODA-Prompt & 86.25 & 4000 & 3.92(72.59$\times$) & 35.17(2.03$\times$) & 26.26$\pm$0.48 & 1.49$\pm$0.04 \\
                \hline
                \addlinespace 
                \multirow{6}{*}{Prompt-Fixed} 
                & L2P & 82.50 & 200 & 0.57(10.56$\times$) & 35.18(2.03$\times$) & 26.00$\pm$0.38 & 1.22$\pm$0.13 \\
                & ADAM-VPT-Deep & 83.20 & 60 & 0.05(1.00$\times$) & 17.28(1.00$\times$) &\textbf{15.58$\pm$0.47} & 0 \\
                & \textbf{PECTP} & \textbf{86.27} & \textbf{60} & \textbf{0.05(1.00$\times$)} & \textbf{17.28(1.00$\times$)} & 22.95$\pm$0.33 & \textbf{0} \\ 
                & PECTP-L2P & 87.82 & 240 & 0.22(4.07$\times$) & - & - & 0 \\ 
                & PECTP-Dual & 88.14 & 600 & 0.54(10.00$\times$) & - & - & 0 \\ 
                & PECTP-CODA & 88.28 & 3600 & 3.24(60.00$\times$) & - & - & 0 \\ 
                \hline
                \addlinespace 
                & Upper Bound & 90.86  & - & 	- &  - & - & 0 \\ 
                \bottomrule
        \end{tabular}}
        \captionsetup{font=small} 
    
        \caption{The trade-off between performance and overheads is crucial when comparing CIFAR performance among three baselines and PECTP. Furthermore,  the results of variants of PECTP (e.g., PECTP-L2P, PECTP-Dual, and PECTP-CODA) with additional learnable parameters are provided to showcase the scalability of PECTP.}
        \label{paper_table_appendix_1}
        \vspace{-3mm}
    \end{table*}
\paragraph{Prompt Number} In existing prompt-based IL methods, the description of `prompt' refers to a set of prompts instead of a single prompt. For example, in L2P, $P_{i} \in \mathbb R^{L_{P}\times D}$, where $L_{P}$ is the number of single prompts, and each $P_{i}$ is stored in a prompt pool. Therefore, for L2P, the total number of prompts is $L_{P} \times$ `the number of prompts'. Furthermore, current IL methods using PTMs follow VPT \cite{jia2022visual} for prompting, which has two variants: VPT-Deep and VPT-shallow. For instance, in DualPrompt \cite{Wang_2022_CVPR}, e-prompts: $e_{i} \in \mathbb R^{L_{e}\times D}$ are inserted into the 3-5 layers of the VIT encoder, while g-prompt: $g_{j} \in \mathbb R^{L_{g}\times D}$ is inserted into the 1-2 layers of the VIT encoder. Therefore, for DualPrompt, the total number of prompts is: `the number of e-prompts' $\times$ $L_{e}\times$ `inserted layers' + `the number of g-prompts' $\times$ $L_{g}\times$ `inserted layers'. From the results in Table \ref{paper_table_appendix_1}, we oberserve that the total number of Prompt Number for L2P, DualPrompt, and CODA-Prompt is around 10, 44, and 72 times that of PECTP, but PECTP outperforms all these methods on $\mathcal{A}_{B}$.
    
\paragraph{Learnable Parameter} Besides Prompt Number another metric is also adopted. Learnable Parameters (LP) are divided into two parts: prompt and key. For L2P, each prompt is matched with a prompt key and jointly trained. This set of prompt keys is used during each inference session, addressing the prompt selection problem. DualPrompt is akin to L2P, with the distinction that it necessitates the addition of a prompt key for each group of `e-prompts,' while `g-prompts' do not require this addition. In CODA-Prompt, the calculation of the `component weight' not only mandates an extra prompt key for each prompt group but also introduces the concept of an `attended-query' to further refine the attention of prompts for different testing samples. Therefore, they also need to add an `attention vector`' for each prompt group. In PECTP, we only need to maintain a single set of prompts, significantly reducing the overhead on learnable parameters. In Table \ref{paper_table_appendix_1}, PECTP achieves the highest performance and the lowest Learnable Parameter after learning the last incremental task. CODA-Prompt exhibit close results to the PECTP. However, the Learnable Parameter required by PECTP are only around 5\% of CODA-Prompt. Furthermore, to demonstrate the scalability of PECTP, we introduce three variants with increased Prompt Number, namely PECTP-L2P, PECTP-Dual, and PECTP-CODA. These three variants have comparable numbers of learnable parameters to L2P, DualPrompt, and CODA-Prompt, respectively, but they outperform the compared methods in terms of performance. Finally, we provide the Upper bound for the CIFAR Inc10 setting, revealing that the gap between PECTP-C and the Upper bound is very small. All the results illustrate that our PECTP framework can strike a good balance between IL performance and memory cost.
\begin{figure}[h]
    \centering
    \begin{subfigure}[b]{0.98\columnwidth}
      \centering
      \includegraphics[width=\textwidth]{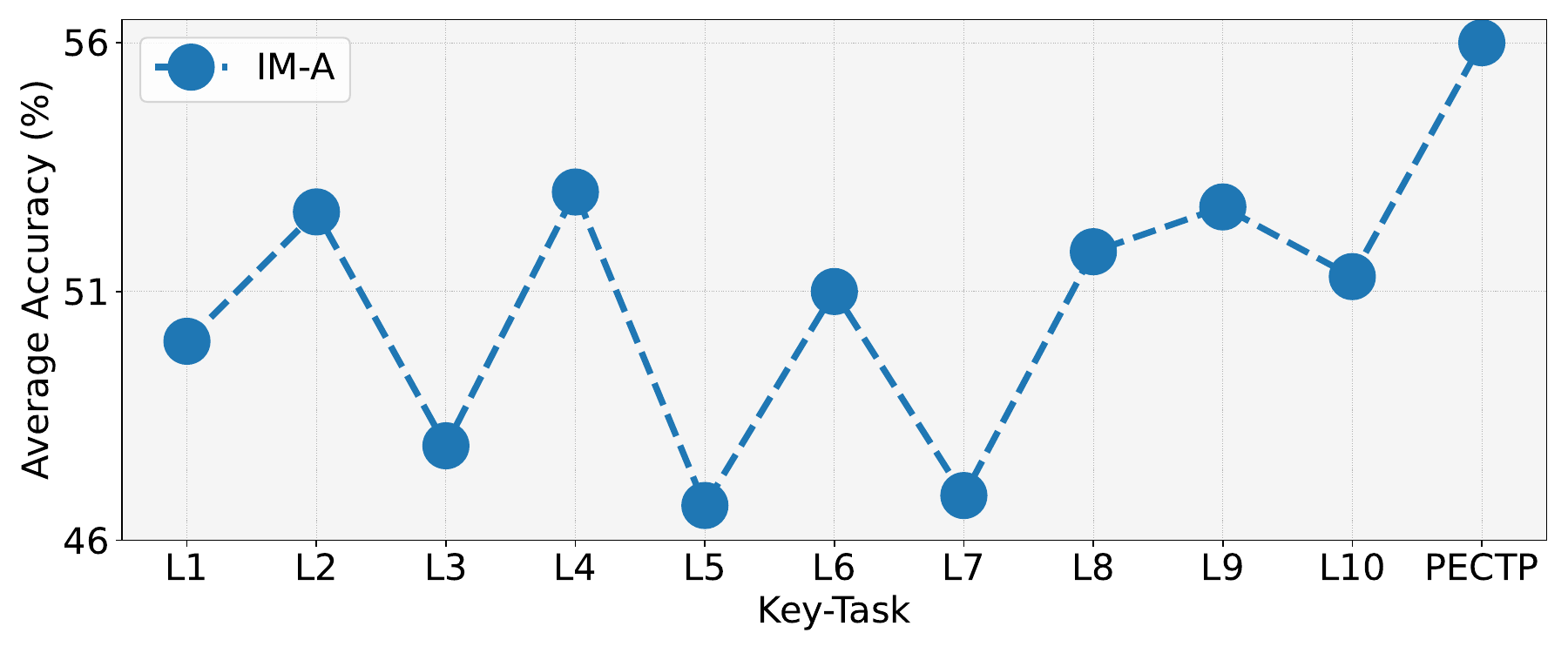}
      \vspace{-5mm}
      \label{lx_pectp_cifar}
    \end{subfigure}
    \vspace{-2mm}
    \begin{subfigure}[b]{0.98\columnwidth}
      \centering
      \includegraphics[width=\textwidth]{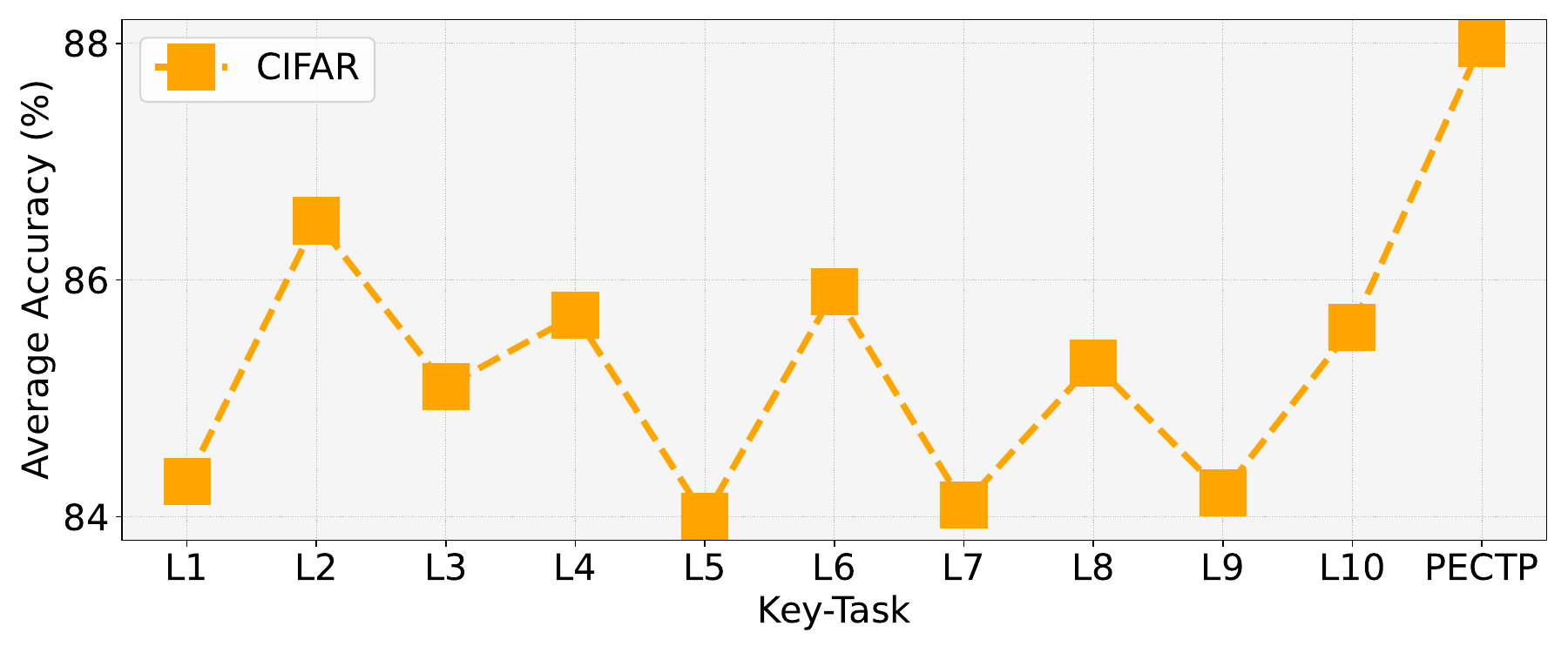}
      \vspace{-5mm}
      \label{lx_pectp_imr}
    \end{subfigure}
    \caption{Results of $\mathcal{A}_{b}$ between L$x$ and PECTP on CIFAR and IM-A.}
        \vspace{-2mm}
    \label{2dataset:key:task}
  \end{figure}
\paragraph{Training Time and Selecting Time}
We conducted a detailed analysis of the time cost for PECTP and baseline methods over one training epoch, denoted as Training Time (TT), and the time spent on prompt selection, denoted as Selecting Time (ST). As shown in Table \ref{paper_table_appendix_1}, except for ADAM-VPT-Deep, PECTP has the shortest TT compared to other methods. However, ADAM-VPT-Deep exhibits a significant performance gap compared to PECTP. Furthermore, since PECTP only needs to maintain a single set of prompts, effectively bypassing the prompt selection problem, the time cost in this step is 0. This further underscores that PECTP, as a parameter-efficient method, not only reduces learnable parameters but also accelerate both training and inference.

\subsection{Detail Analysis}
\paragraph{Real Gains over PTMs} 
It faces a certain degree of data leakage because PTM is pre-trained on ImageNet, and there is partial data-overlap between the pre-training data and the training data for downstream incremental tasks. Following \cite{zhou2023revisiting}, we introduce a novel baseline for Prompt-Based IL methods, referred to SimpleCIL. Specifically, SimpleCIL performs directly on incremental tasks without any prompting or fine-tuning. The results can be found in the seventh row of Table \ref{tab:sub1} and the forth row of Table \ref{tab:sub2}, demonstrating that PECTP consistently outperforms SimpleCIL across seven benchmarks, especially on those without data-overlap (e.g., IM-A, ObjNet, Omin, VTAB).

\paragraph{Gains on each Task} 

Figure \ref{gain_on_each_cifar} and Figure \ref{gain_on_each_ima} present detailed accuracy for each task. Here, we provide a comparison between PECTP and ADAM-VPT-Deep under two experimental settings. Since PECTP is trained on each incremental task, it consistently surpasses ADAM-VPT-Deep across all incremental tasks. Additionally, we present PlainCIL, which imposes no constraints when learning new tasks. The results further demonstrates that the proposed PRM and HRM effectively reduce forgetting of old knowledge when learning new tasks.

\paragraph{Effectiveness of Cross-Task Prompt} 
\label{Cross-Task Prompt vs. Key-Task Prompt}

Compared to prompt-fixed methods that employ key-task prompts, our PECTP uses a single set of prompts that learns across each incremental task. In prompt-fixed methods, the key-task is typically the first task. Therefore, by deliberately adjusting the order of the sequential tasks, we can select different tasks as the key-task. For simplicity, we denote the key-task prompts learned on the $D_{x}$ task as L$x$. As shown in Figure \ref{2dataset:key:task}, the accuracy of PECTP consistently surpasses that of L$x$. 

Furthermore, we visualize the extracted features using T-SNE. As shown in Figure \ref{paper:tsne}, features extracted through prompt-fixed methods are insufficient, resulting in a fuzzy classification boundary (blue box). Meanwhile, prompt-extending methods, compared to prompt-fixed methods, attempt to better differentiate class $6,7,9$ from class $2,3,5$. However, when making finer distinctions within class $6,7,9$, noticeable feature mixing occurs. In contrast, PECTP maintains a clear classification boundary.

\section{Conclusion}
In this paper, we propose the \uline{P}arameter-\uline{E}fficient \uline{C}ross-\uline{T}ask \uline{P}rompt (i.e., PECTP) for Rehearsal-Free and Memory-Constrained Incremental Learning (RFMCIL). We first conduct a detailed analysis of the prompt-extending and prompt-fixed prompt-based methods. To address the performance concerns of prompt-fixed methods, we propose training the introduced prompt not only on the first (or key-task) incremental task but also on each task to continuously acquire new knowledge. To address the computational overhead and parameter storage issues faced by prompt-extending methods, we propose using a single but efficient set of prompts, thereby avoiding the need to maintain a prompt pool and select from multiple sets of prompts. PECTP utilizes a PRM to restrict parameter evolution of cross-task prompts from the OPG and IPG, which can effectively preserve the learned knowledge of the prompts after learning the new incremental task. Additionally, we propose a HRM to facilitate the generalization of the cross-task prompts. We perform extensive evaluations of our method and other prompt-based methods, demonstrating the effectiveness of our approach. The proposed PECTP, due to its use of a single set of prompts, is inherently limited in performance, especially when faced with downstream tasks that are significantly different, such as those that span across domains and modalities, or have extremely imbalanced data distributions. A simple idea to flexibly combine prompt-extending methods with PECTP and dynamically select between them will be the direction of our future exploration.

\normalem
\bibliographystyle{IEEEtran}
\bibliography{Weixia}{}

\end{document}